\documentclass[sn-mathphys,Numbered]{sn-jnl}

\usepackage{graphicx}%
\usepackage{multirow}%
\usepackage{amsmath,amssymb,amsfonts}%
\usepackage{amsthm}%
\usepackage{mathrsfs}%
\usepackage[title]{appendix}%
\usepackage{xcolor}%
\usepackage{textcomp}%
\usepackage{manyfoot}%
\usepackage{booktabs}%
\usepackage{algorithm}%
\usepackage{algorithmicx}%
\usepackage{algpseudocode}%
\usepackage{listings}%
\usepackage{subcaption}
\usepackage{float}
\usepackage{color}
\usepackage{amsmath,amssymb,graphicx}
\usepackage{todonotes}
\usepackage{enumitem}
\usepackage{multirow}

\usepackage{algorithm}%
\usepackage{algorithmicx}%
\usepackage{algpseudocode}%

\definecolor{codegreen}{rgb}{0,0.6,0}
\definecolor{codegray}{rgb}{0.5,0.5,0.5}
\definecolor{codepurple}{rgb}{0.58,0,0.82}
\definecolor{backcolour}{rgb}{0.95,0.95,0.92}
\lstdefinestyle{mystyle}{
    backgroundcolor=\color{backcolour},
    commentstyle=\color{codegreen},
    keywordstyle=\color{codepurple},
    numberstyle=\tiny\color{white},
    stringstyle=\color{blue},
    basicstyle=\ttfamily\small,
    breakatwhitespace=false,
    breaklines=true,
    captionpos=b,
    keepspaces=true,
    numbers=left,
    numbersep=1pt,
    showspaces=false,
    showstringspaces=false,
    showtabs=false,
    tabsize=2,
    language=Python
}
\definecolor{mypink2}{RGB}{219, 48, 122}

\newcommand\ri[1]{{#1}}

\begin{document}

\title[Calibrated Explanations for Regression]{Calibrated Explanations for Regression}


\author*[1]{\fnm{Tuwe} \sur{Löfström}}\email{tuwe.lofstrom@ju.se}

\author[1]{\fnm{Helena} \sur{Löfström}}\email{helena.lofstrom@ju.se}

\author[1]{\fnm{Ulf} \sur{Johansson}}\email{ulf.johansson@ju.se}

\author[1]{\fnm{Cecilia} \sur{Sönströd}}\email{cecilia.sonstrod@ju.se}

\author[1]{\fnm{Rudy} \sur{Matela}}\email{rudy.matela@ju.se}

\affil*[1]{\orgdiv{Jönköping AI Lab (JAIL), Department of Computing}, \orgname{Jönköping University}, \orgaddress{\street{Box 1026}, \city{Jönköping}, \postcode{55111}, \country{Sweden}}}




\abstract{Artificial Intelligence (AI) is often an integral part of modern decision support systems. The best-performing predictive models used in AI-based \ri{decision support systems} lack transparency. Explainable Artificial Intelligence (XAI) aims to create AI systems that can explain their rationale to human users. Local explanations in XAI can provide information about the causes of individual predictions in terms of feature importance. However, a critical drawback of existing local explanation methods is their inability to quantify the uncertainty associated with a feature's importance. This paper introduces an extension of a feature importance explanation method, Calibrated Explanations, previously only supporting classification, with support for standard regression and probabilistic regression, i.e., the probability that the target is above an arbitrary threshold. The extension for regression keeps all the benefits of \ri{Calibrated Explanations}, such as calibration of the prediction from the underlying model with confidence intervals, uncertainty quantification of feature importance, and allows both factual and counterfactual explanations. \ri{Calibrated Explanations} for standard regression provides fast, reliable, stable, and robust explanations. \ri{Calibrated Explanations} for probabilistic regression provides an entirely new way of creating probabilistic explanations from any ordinary regression model, allowing dynamic selection of thresholds. 
The method is model agnostic with easily understood conditional rules. An implementation in Python is freely available on GitHub and for installation using both \texttt{pip} and \texttt{conda}, making the results in this paper easily replicable.}

\keywords{Explainable AI, Feature Importance, Calibrated Explanations, Uncertainty Quantification, Regression, Probabilistic Regression, Counterfactual Explanations, Conformal Predictive Systems}



\maketitle

\section{Introduction}\label{sec1}
\noindent In recent times, Decision Support Systems in various domains such as retail, sport, or defence have been incorporating Artificial Intelligence (AI) extensively \citep{zhou2021evaluating}. However, the predictive models used in AI-based \ri{Decision Support Systems} generally lack transparency and only provide probable results \citep{DavidGunning2017,Ribeiro2016_kdd}. This can result in misuse (when users rely on it excessively) or disuse (when users do not rely on it enough) \citep{alvarado2014reliance,buccinca2020proxy}.

The lack of transparency has led to the development of eXplainable Artificial Intelligence (XAI), which aims to create AI systems capable of explaining their reasoning to human users. The goal of explanations is to support users in identifying incorrect predictions, especially in critical areas such as medical diagnosis \citep{gunning19}. An explanation provided by XAI should highlight the underlying model's strengths and weaknesses and provide insight into how it will perform in the future \citep{DavidGunning2017, dimanov2020you}. 

Regarding explanations in XAI, there are two types: local and global. Local explanations focus on the reasons behind individual predictions, while global explanations provide information about the entire model \citep{guidotti2018survey,moradi2021post,Martens14}. Despite the apparent strength stemming from the possibility of providing explanations for each instance, local explanations typically have some drawbacks. For example, they can be instable, meaning that the same model and instance may result in different explanations, or they can lack robustness, meaning that minor differences in the instance can lead to significantly different explanations \citep{slack2021reliable, rahnama2019study}. Instability and lack of robustness create issues when evaluating the quality of the explanations. Metrics like fidelity, which measure how well an explanation captures the behaviour of the underlying model, do not give an accurate picture of explanation quality since they depend heavily on the details of the explanation method \citep{slack2021reliable, moradi2021post, hoffman2018metrics, carvalho2019machine, adadi2018peeking, wang2019Designing, mueller19, agarwal2022openxai}. Furthermore, even the best explanation techniques offer limited insight into model uncertainty and reliability. Recent research has emphasized uncertainty estimation's role in enhancing the transparency of underlying models \citep{bhatt2021uncertainty, slack2021reliable}. Although achieving well-calibrated uncertainty has been underscored as a critical factor in fostering transparent decision-making, \ri{\citet{bhatt2021uncertainty} point out} the challenges and complexities of obtaining accurately calibrated uncertainty estimates for complex problems. Moreover, as indicated by \ri{\citet{slack2021reliable}}, the focus has predominantly leaned towards adopting a well-calibrated underlying model (such as Bayesian) rather than relying on calibration techniques. 

The probability estimate that most classifiers output is commonly used as an indicator of the likelihood of each class in local explanation methods for classification. However, it is widely recognized that these classifiers are often poorly calibrated, resulting in probability estimates that do not faithfully represent the actual probability of correctness \cite{vovk2015cross}. Specialized calibration techniques such as Platt Scaling \citep{platt1999probabilistic} and Venn-Abers (VA) \citep{vovk2012vennabers} have been proposed to tackle these shortcomings. The VA method generates a probability range associated with each prediction, which can be refined into a properly calibrated probability estimate utilizing regularisation. 

When employing the VA approach for decision-making, it is essential to recognize that the technique provides intervals for the positive class. These intervals quantify the uncertainty within the probability estimate, offering valuable insights from an explanatory standpoint. The breadth of the interval directly corresponds to the model's level of uncertainty, with a narrower interval signifying more confidence in the probability estimate. In comparison, a broader interval indicates more substantial uncertainty in said estimates. The uncertainty information can be extended to the features, given that the feature weights are informed by the prediction's probability estimate. Being able to quantify the uncertainty of feature weights can improve the quality and usefulness of explanations in XAI. Recently, a local explanation method, Calibrated Explanations, utilizing the intervals provided by VA to estimate feature uncertainty was introduced for classification \citep{calibrated-explanations}.

\ri{In recent years, conformal prediction has increasingly been integrated into research about XAI methods, although not focusing on the uncertainty aspect per se. The focus has primarily been on interpretable models \cite{johanssoncopa19a}, increasing the fidelity between model and explanations \cite{altmeyer2024faithful}, lowering the computational cost \cite{alkhatib2023approximating} and explaining reject options \cite{10022139,artelt2022model,artelt2023not}. Explaining reject options has been defined as an explanation of the uncertainty integral in taking a decision.}

Existing explanation methods most commonly focus on explaining decisions from classifiers, despite the fact that regression is widely used in highly critical situations. Due to the lack of specialized explanation techniques for regression, applying methods designed for classification on regression problems is not unusual, highlighting the need for well-founded explanation methods for regression \citep{letzgus2022toward}. 

The aim of this study is to propose an explanation method - with the same possibility of quantifying the uncertainty of feature weights that \ri{Calibrated Explanations provides, through VA,} for classification - for a regression context. 
The conformal prediction framework \citep{vovk2005algorithmic} provides several different techniques for quantifying uncertainty in a regression context. In this paper, the Conformal Predictive Systems (CPSs) technique \citep{VovkSMX19} for uncertainty estimation is used in \ri{Calibrated Explanations} to allow creation of calibrated explanations with uncertainty estimation for regression. CPSs is not only a very flexible technique, providing a rich set of tools to be used for uncertainty quantification, but it also allows for estimating the probability that the target is above any user-defined threshold. Based on this, a new form of probabilistic explanation for regression is also proposed in this paper. These approaches are user-friendly and model-agnostic, making them easy to use and applicable to diverse underlying models. 


In summary, this paper introduces extensions of \ri{Calibrated Explanations} aimed at regression, with the following characteristics: 
\begin{itemize}
    \item Fast, reliable, stable and robust feature importance explanations for regression. 
    \item Calibration of the predictions from the underlying model through the application of CPSs.
    \item \ri{Explanations with a}rbitrary forms of uncertainty quantification of the predictions from the underlying model and the feature importance weights through querying of the conformal predictive distribution (CPD) derived from the CPS.
    \item Possibility of creating explanations of the probability that the prediction exceeds a user-defined threshold\ri{, with uncertainty quantification}.
    \item Rules with straightforward interpretation in relation to the feature values and the target.
    \item Possibility to generate counterfactual rules with uncertainty quantification of the expected predictions (or probability of exceeding a threshold).
    \item \ri{Conjunctive rules can be created, conveying feature importance for the interaction of included features.}
    \item Distribution as an open source Python package, making the proposed techniques easily accessible for both scientific and industrial purposes. 
\end{itemize}

\section{Background}
\subsection{Post-Hoc Explanation Methods} \label{sec:Expl_meth}
\noindent The research area of XAI can be broadly categorized into two main types: developing inherently interpretable and transparent models and utilizing \textit{post-hoc methods} to explain opaque models. Post-hoc explanation techniques seek to construct simplified and interpretable models that reveal the relationship between feature values and the model's predictions. These explanations, which can be either local or global, often leverage visual aids such as pixel representations, feature importance plots, or word clouds, emphasizing the features, pixels, or words accountable for causing the model's predictions \citep{molnar2020interpretable, moradi2021post}.

Two distinct approaches of explanations exist: \textit{factual} explanations, where a feature value directly influences the prediction outcome, and \textit{counterfactual} explanations, which explore the potential impact on predictions when altering a feature's values \citep{mothilal2020explaining, guidotti2022counterfactual, wachter2017counterfactual}. Importantly, counterfactual explanations are intrinsically local. They are particularly human-friendly, mirroring how human reasoning operates \citep{molnar2020interpretable}.

\subsection{Essential Characteristics of Explanations}
\noindent Creating high-quality explanations in XAI requires a multidisciplinary approach that draws knowledge from both the Human-Computer Interaction and the Machine Learning fields. The quality of an explanation method depends on the goals it addresses, which may vary. For instance, assessing how users appreciate the explanation interface differs from evaluating if the explanation accurately mirrors the underlying model \citep{Lofstrom2022}. However, specific characteristics are universally desirable for post-hoc explanation methods. It is crucial that an explanation method accurately reflects the underlying model, which is closely related to the concept that an explanation method should have a high level of fidelity to the underlying model \citep{slack2021reliable}. Therefore, a reliable explanation must have feature weights that correspond accurately to the actual impact on the estimates to correctly reflect the model's behavior \citep{bhatt2021uncertainty}. 

Stability and robustness are two additional critical features of explanation methods \citep{dimanov2020you, agarwal2022openxai, alvarez2018robustness}. Stability refers to the consistency of the explanations \citep{slack2021reliable, carvalho2019machine}; the same instance and model should produce identical explanations across multiple runs. On the other hand, robustness refers to the ability of an explanation method to produce consistent results even when an instance undergoes small perturbations \citep{dimanov2020you} or other circumstances change. Therefore, the essential characteristics of an explanation method in XAI are that it should be reliable, stable, and robust.

\subsection{Explanations for classification and regression}
\noindent Distinguishing between explanations for classification and regression lies in the nature of the insights they offer. In classification, the task involves predicting the specific class an instance belongs to from a set of predefined classes. The accompanying probability estimates reflect the model's confidence level for each class. Various explanation techniques have been developed for classifiers to clarify the rationale behind the class predictions. Notable methods include SHAP \citep{lundberg2017unified}, LIME \citep{Ribeiro2016_kdd}, and Anchor \citep{ribeiro2018anchors}. These techniques delve into the factors that contribute to the assignment of a particular class label. Typically, the explanations leverage the concept of feature importance, e.g., words in textual data or pixels in images.

In regression, the paradigm shifts as there are no predetermined classes or categorical values. Instead, each instance is associated with a numerical value, and the prediction strives to approximate this value. Consequently, explanations for regression models cannot rely on the framework of predefined classes. Nevertheless, explanation techniques designed for classifiers, as mentioned above, can often be applied to regression problems, provided these methods concentrate on attributing features to the predicted instance's output.

\subsection{Venn-Abers predictors}\label{VennAbers}
\noindent Probabilistic predictors offer class labels and associated probability distributions. Validating these predictions is challenging, but calibration focuses on aligning predicted and observed probabilities \citep{vovk2005algorithmic}. The goal is well-calibrated models where predicted probabilities match actual accuracy. Venn predictors \citep{vovk2004selfcalibrating} produce multi-probabilistic predictions, converted to confidence-based probability intervals.

Inductive Venn prediction \citep{Lambrou2015} involves a Venn taxonomy, categorizing calibration data for probability estimation. Within each category, the estimated probability for test instances falling into a category is the relative frequency of each class label among all calibration instances in that category. 

Venn-Abers predictors (VA) \citep{vovk2012vennabers} offer automated taxonomy optimization via isotonic regression, thus introducing dynamic probability intervals. A two-class scoring classifier assigns a prediction score $s_i$ to an object $x_i$. A higher score implies higher belief in the positive class. In order to calibrate a model, some data must be set aside and used as a calibration set \ri{when using inductive VA predictors}. Consequently, split the training set $\{z_1, \dots, z_i, \dots, z_{n}\}$, with objects $x_i$ and labels $y_i$, into a proper training set $Z_T$ and a calibration set $\{z_{1}, \dots, z_q\}$\footnote{As we assume random ordering, the calibration set is indexed $1,\dots,q$ rather than $|Z_T|+1,\dots,n$, for indexing convenience.}. Train a scoring classifier on $Z_T$ to compute $s$ for $\{x_{1},\dots,x_q,x\}$\ri{, where $x$ is the object of the test instance $z$\footnote{The index $n+1$ is dropped for indexing convenience whenever referring to the test instance (like $z_{n+1}, x_{n+1}$ or $y_{n+1}$) or values dependent on the test instance (like $s_{n+1}$).}}. Inductive VA prediction follows these steps:

\begin{enumerate}
    \item Derive isotonic calibrators $g_0$ and $g_1$ using $\{\{s_1,y_1\},\dots,\{s_q,y_q\},\{s,y=0\}\}$ and $\{\{s_1,y_1\},\dots,\{s_q,y_q\},\{s,y=1\}\}$, respectively.
    \item The probability interval for $y=1$ is $[g_0(s),g_1(s)]$ (henceforth referred to as $[\mathcal{P}_l,\mathcal{P}_h]$, representing the lower and upper bounds of the interval).
    \item Obtain a regularized probability estimate for $y=1$ using the recommendation by \cite{vovk2012vennabers}:
\begin{equation*}
\mathcal{P}=\frac{\mathcal{P}_h}{1-\mathcal{P}_l+\mathcal{P}_h}
\end{equation*}
\end{enumerate}

\ri{Since the class label of the test instance must be either positive or negative in binary classification, and the lower and upper bounds are the relative frequencies calculated from the calibration set (including the test instance with the positive or negative label assigned), one of them must be the correctly calibrated probability estimate. Thus, the probability interval is well-calibrated provided the data is exchangeable.}

In summary, VA produces a calibrated (regularized) probability estimate $\mathcal{P}$ together with a probability interval with a lower and upper bound $[\mathcal{P}_l,\mathcal{P}_h]$.

\subsection{Calibrated Explanations for Classification}
\label{CE}
\noindent Below is an introduction to \ri{Calibrated Explanations for classification} \citep{calibrated-explanations}, which provides the foundation to this paper's contribution. 
In the following descriptions, a \textit{factual explanation} is composed of a \textit{calibrated prediction} from the underlying model accompanied by an \textit{uncertainty interval} and a collection of \textit{factual feature rules}, each composed of a \textit{feature weight with an uncertainty interval} and a \textit{factual condition}, covering that feature's instance value. \textit{Counterfactual explanations} only contain a collection of \textit{counterfactual feature rules}, each composed of a \textit{prediction estimate with an uncertainty interval} and a \textit{counterfactual condition}, covering alternative instance values for the feature. The prediction estimate represents a probability estimate for classification, whereas for regression, the prediction estimate will be expressed as a potential prediction. 

\subsubsection{Factual \ri{Calibrated} Explanations for Classification} \label{CEC}

\noindent \ri{Calibrated Explanations} is applied to an underlying model with the intention of explaining its predictions of individual instances using rules conveying feature importances. 
\ri{The following is a high-level description of how Calibrated Explanations for classification works, following the original description in \citep{calibrated-explanations} closely:}

\ri{Let us assume that a scoring classifier, trained using the proper training set $Z_T$, exists for which a local explanation for test object $x$ is wanted. Use VA as a calibrator and calibrate the underlying model for $x$ to get the probability interval $[\mathcal{P}_l, \mathcal{P}_h]$ and the calibrated probability estimate $\mathcal{P}$. For each feature $f$, use the calibrator to estimate probability intervals ($[\mathcal{P}'_{l.f}, \mathcal{P}'_{h.f}]$) and calibrated probability estimates ($\mathcal{P}'_{f}$) for slightly perturbed versions of object $x$, changing one feature at a time in a systematic way (see the detailed description below). To get the feature weight (and uncertainty interval) for feature $f$, calculate the difference between $\mathcal{P}$ to the average of all $\mathcal{P}'_{f}$ (and $[\mathcal{P}'_{l.f}, \mathcal{P}'_{h.f}]$)\footnote{Exclude $\mathcal{P}$ (and $[\mathcal{P}_l, \mathcal{P}_h]$), i.e., the calibrator's results on $x$. \ri{There may be alternatives to how we compute these weights (see Section~\ref{future-work}).}}: 
\begin{equation}
    w_f = \mathcal{P} - \frac{1}{|V_f|-1}\sum \mathcal{P}'_{f},
    \label{eq:w_f}
\end{equation}
\begin{equation}
    w_l^f = \mathcal{P} - \frac{1}{|V_f|-1}\sum \mathcal{P}'_{l.f}, 
    \label{eq:w_0^f}
\end{equation}
\begin{equation}
    w_h^f = \mathcal{P} - \frac{1}{|V_f|-1}\sum \mathcal{P}'_{h.f},
    \label{eq:w_1^f}
\end{equation}
where $|V_f|-1$ is the number of perturbed values. }
        
\ri{The feature weight is exactly defined to be the difference between the calibrated probability estimate on the original test object $x$ and the estimated (average) calibrated probability estimate achieved on the perturbed versions of $x$. The upper and lower bounds are defined analogously using the probability intervals from the perturbed versions of $x$. As long as the same test object, underlying model and calibration set is used, the resulting explanation will also be the same.}

\ri{More formally, the following steps are pursued to achieve a factual explanation for a test object $x$:
\begin{enumerate}
  \setlength{\itemsep}{1pt}
  \setlength{\parskip}{0pt}
  \setlength{\parsep}{0pt}
    \item Use the calibrator to get the probability interval $[\mathcal{P}_l, \mathcal{P}_h]$ and the calibrated probability estimate $\mathcal{P}$ for $x$. 
    \item Separate all features into categorical features $C$ and numerical features $N$. Define a discretizer for numerical features that define thresholds and smaller-or-equal- or greater-than-conditions $(\leq, >)$ for these features\footnote{This is done using a binary subclass of the \texttt{EntropyDiscretizer} class in LIME.}. 
    \item For each feature $f\in C$: \label{step3}
        \begin{enumerate}
              \setlength{\itemsep}{1pt}
              \setlength{\parskip}{0pt}
              \setlength{\parsep}{0pt}
            \item Iterate over all possible categorical values $v\in V_f$ and create a perturbed instance exchanging the feature value of $x_f$ with one value at a time, creating a perturbed instance $x'_{f}=v$. 
            \item Calculate and store the probability intervals $[\mathcal{P}'_{l.f}, \mathcal{P}'_{h.f}]$ and the calibrated probability estimate $\mathcal{P}'_{f}$ for the perturbed instance. Calculate the weights using equations~(\ref{eq:w_f}), (\ref{eq:w_0^f}), and (\ref{eq:w_1^f}). 
            \item Define a factual condition using the the feature $f$, the value $v$ and the identity condition $(=)$. \label{cat_cond}
        \end{enumerate}    
    \item For each feature $f\in N$: \label{step4}
        \begin{enumerate}
              \setlength{\itemsep}{1pt}
              \setlength{\parskip}{0pt}
              \setlength{\parsep}{0pt}
            \item Use the thresholds of the discretizer to identify the closest lower or upper threshold $t$ surrounding the feature value of $x_f$. Divide all possible feature values in the calibration set for feature $f$ into two groups $V_f$ separated by $t$. \label{find_threshold}
            \item Within each group, percentile values $pv$ representing the $25^{th}$, $50^{th}$ and $75^{th}$ percentiles are extracted. 
                Iterate over the values in $pv$ and create a perturbed instance exchanging the feature value of $x_f$ with one value at a time, creating a perturbed instance $x'_{f_{pv}}$. Calculate and store the probability intervals $[\mathcal{P}'_{l.f_{pv}}, \mathcal{P}'_{h.f_{pv}}]$ and the calibrated probability estimate $\mathcal{P}'_{f_{pv}}$ for the perturbed instance. 
                Average over all perturbed instances within the group, creating a probability interval $[\mathcal{P}'_{l.f}, \mathcal{P}'_{h.f}]$ and the calibrated probability estimate $\mathcal{P}'_{f}$ for each group. Calculate the weights using equations~(\ref{eq:w_f}), (\ref{eq:w_0^f}), and (\ref{eq:w_1^f}). 
            \item Define a factual condition using threshold $t$ and feature $f$. The $\leq \text{or} >$ condition is used so that the factual condition covers the instance value. \label{num_cond}
        \end{enumerate} 
\end{enumerate} }

\subsubsection{Counterfactual \ri{Calibrated} Explanations for Classification}
\noindent When creating factual explanations, the calibrator's results from perturbed instances are averaged to calculate feature importance and uncertainty intervals for each feature. When generating counterfactual rules, the calibrator's results for perturbed instances are instead used to form counterfactual rules. For categorical features, one counterfactual rule is created for each alternative categorical value, and for numerical features, (up to) two rules, representing $\leq$-rules and $>$-rules, \ri{can} be created\ri{\footnote{All existing Discretizers in LIME can be used for counterfactual explanations. The \texttt{EntropyDiscretizer} class is used by default for counterfactual classification.}.} Each feature rule's expected probability interval is already established as \ri{$[\mathcal{P}'_{l.f}, \mathcal{P}'_{h.f}]$}, following the \ri{Calibrated Explanations} process in \ri{steps~\ref{step3} and~\ref{step4} above}, defining one feature rule for each alternative instance value. The condition will be similar as in \ri{steps~\ref{step3} and~\ref{step4} above}, but for the alternative instance value $v$. Equation~(\ref{eq:w_f})'s feature weights are mainly employed to sort counterfactual rules by impact. The calibrated probability estimate \ri{$\mathcal{P}'_{f}$} is normally neglected in counterfactual rules for classification but is calculated and can be used.

\subsubsection{\ri{Conjunctive Calibrated Explanations}}
Each individual rule only conveys the contribution of an individual feature. To counteract this shortcoming, conjoined rules can be derived to estimate the joint contribution between combinations of features. This is done separately from the generation of the feature rules, by combining the \ri{established} feature rules. \ri{For each combination of existing rules, new perturbed instances are created by applying the already established feature rule conditions, limiting the search space of conjunctions to consider to the most important existing rules. Calibration is performed following the same logic as for single feature perturbed instances, making it possible to get well-calibrated conjunctive rules taking feature interaction into account.} 

\section{Calibrated Explanations for Regression}
\noindent The basic idea in \ri{Calibrated Explanations for classification} is that each factual and counterfactual explanation is derived using three calibrated values: The calibrated probability and the probability interval represented by the lower and upper bound. 

For regression, there are two natural use cases, where the obvious one is predicting the continuous target value directly, i.e., standard regression. Another use case is to instead predict the probability of the target being below (or above) a given threshold, basically viewing the problem as a binary classification problem. 

Conformal Predictive Systems (CPSs) produce Conformal Predictive Distributions (CPDs), as mentioned in the introduction. CPDs are cumulative distribution functions which can be used for various purposes, such as deriving prediction intervals for specified confidence levels or obtaining the probability of the true target falling below (or above) any threshold. 

\subsection{Conformal Predictive Systems}  \label{sec:cps}\label{sec:CR} 

\noindent Conformal prediction \citep{vovk2005algorithmic} offer predictive confidence by generating prediction regions, which include the true target with a specified probability. These regions are sets of class labels for classification or prediction intervals for regression.

Errors arise when the true target falls outside the region, yet conformal predictors are automatically valid under exchangeability, yielding an error rate of $\varepsilon$ over time. Thus, the key evaluation criterion is efficiency, gauged by the region's size and sharpness for greater insight. Conformal regressors (CRs), specifically an \textit{inductive (split) CR} \citep{papadopoulos2002inductive}, follows these steps:
\begin{enumerate}
    \item Divide the data into a proper training set $Z_T$ and a calibration set $\{z_{1}, \dots, z_q\}$.
    \item Fit an underlying regression model $h$ to $Z_T$.
    \item Define nonconformity as the absolute error $\left| y_i - h(x_i) \right|$.
    \item Compute nonconformity scores for $\{z_{1}, \dots, z_q\}$ and sort them in descending order to obtain $\alpha_1\leq\alpha_2\leq...\leq\alpha_q$.
    \item Assign an $\varepsilon$, e.g., 0.01, 0.05, or 0.1.
    \item Calculate the $(1-\varepsilon)$-percentile nonconformity score, $\alpha_s$, where index $s = \left\lfloor\varepsilon(q+1)\right\rfloor$.
    \item For a new instance $x_i$, the prediction interval is $h(x_i) \pm \alpha_s$.
\end{enumerate}
To individualize intervals, the \textit{normalized nonconformity function} \citep{papadopoulosnormalized} augments nonconformity with $\sigma_i$ and $\beta$. These adapt intervals based on predicted difficulty $\sigma_i$ for each $y_i$. Normalized nonconformity is $\frac{\left|y_i - h(x_i)\right|}{\sigma_i+\beta}$, and the interval is $h(x_i) \pm \alpha_s(\sigma_i+\beta)$. This approach yields individualized prediction intervals, accommodating prediction difficulty and enhancing region informativeness.

The process of creating (normalized) inductive CPSs closely resembles the formation of inductive CRs \citep{VovkSMX19}. The primary distinction lies in calculating nonconformity scores using actual errors, defined as:
\begin{equation}
    f\left(z_i\right) = y_i - h\left(x_i\right),
    \label{eqn:abs-error-cps}
\end{equation}
or normalized errors:
\begin{equation}
    f\left(z_i\right) = \frac{y_i - h\left(x_i\right)}{\sigma_i  + \beta},
    \label{eqn:abs-error-norm-cps}
\end{equation}
where $\sigma_i$, $x_i$, and $\beta$ retain their prior definitions. The prediction for a test instance $x_i$ (potentially with an estimated difficulty $\sigma_i$) then becomes the following CPD:

\begin{equation}
    \displaystyle
    \mathcal{Q}(y) = 
        \begin{cases}
            \textstyle \frac{i+\tau}{q+1}, \textrm{ if } y\in\left(\mathcal{C}_{(i)},\mathcal{C}_{(i+1)}\right), & \textrm{for } i \in \{0,...,q\}\\
            \textstyle \frac{i'-1+(i''-i'+2)\tau}{q+1}, \textrm{if } y = \mathcal{C}_{(i)}, & \textrm{for } i\in \{1,...,q\}
        \end{cases}  
    \label{eqn:cps-q}
\end{equation}
where $\mathcal{C}_{(1)}, \ldots, \mathcal{C}_{(q)}$ are obtained from the calibration scores $\alpha_1, \ldots, \alpha_q$, sorted in increasing order:
\begin{equation*}
    \mathcal{C}_{(i)} = h\left(x\right)+\alpha_i
\end{equation*}
or, when using normalization: 
\begin{equation*}
    \mathcal{C}_{(i)} = h\left(x\right)+\sigma\alpha_i
\end{equation*}
with $\mathcal{C}_{(0)}=-\infty$ and $\mathcal{C}_{(q+1)}=\infty$. $\tau$ is sampled from the uniform distribution $\mathcal{U}(0,1)$ and its role is to allow the $\mathcal{P}$-values of target values to be uniformly distributed. $i''$ is the highest index such that $y = \mathcal{C}_{(i'')}$, while $i'$ is the lowest index such that $y = \mathcal{C}_{(i')}$ (in case of ties). For a specific value $y$, the function returns the estimated probability $\mathcal{P}(\mathcal{Y} \leq y)$, where $\mathcal{Y}$ is a random variable corresponding to the true target. 

Given a CPD: 
\begin{itemize}
    \item A two-sided prediction interval for a chosen significance level $\varepsilon$ can be obtained by $[\mathcal{C}_{\lfloor (\varepsilon/2)(q+1) \rfloor}, \mathcal{C}_{\lceil (1-\varepsilon/2)(q+1) \rceil}]$. \ri{Obviously, the interval does not have to be symmetric as long as the covered range of percentiles are $1-\varepsilon$.} 
    \item One-sided prediction intervals can be obtained by $[\mathcal{C}_{\lfloor \varepsilon(q+1) \rfloor}, \infty]$ for a lower-bounded interval, and by $[-\infty, \mathcal{C}_{\lceil (1-\varepsilon)(q+1) \rceil}]$ for an upper-bounded interval. 
    \item Similarly, a point prediction corresponding to the median of the distribution can be obtained by $(\mathcal{C}_{\lceil 0.5(q+1) \rceil}+\mathcal{C}_{\lfloor 0.5(q+1) \rfloor})/2$. \ri{Since the median is an unbiased midpoint in the distribution measured on the calibration set, t}he median prediction can be seen as a calibration of the underlying models prediction. 
    Unless the model is biased, the median will tend to be very close to the prediction of the underlying model.
\end{itemize}
\begin{figure}
    \centering
    \includegraphics[width=0.9\textwidth, trim={2.7cm 1cm 3cm 2cm},clip]{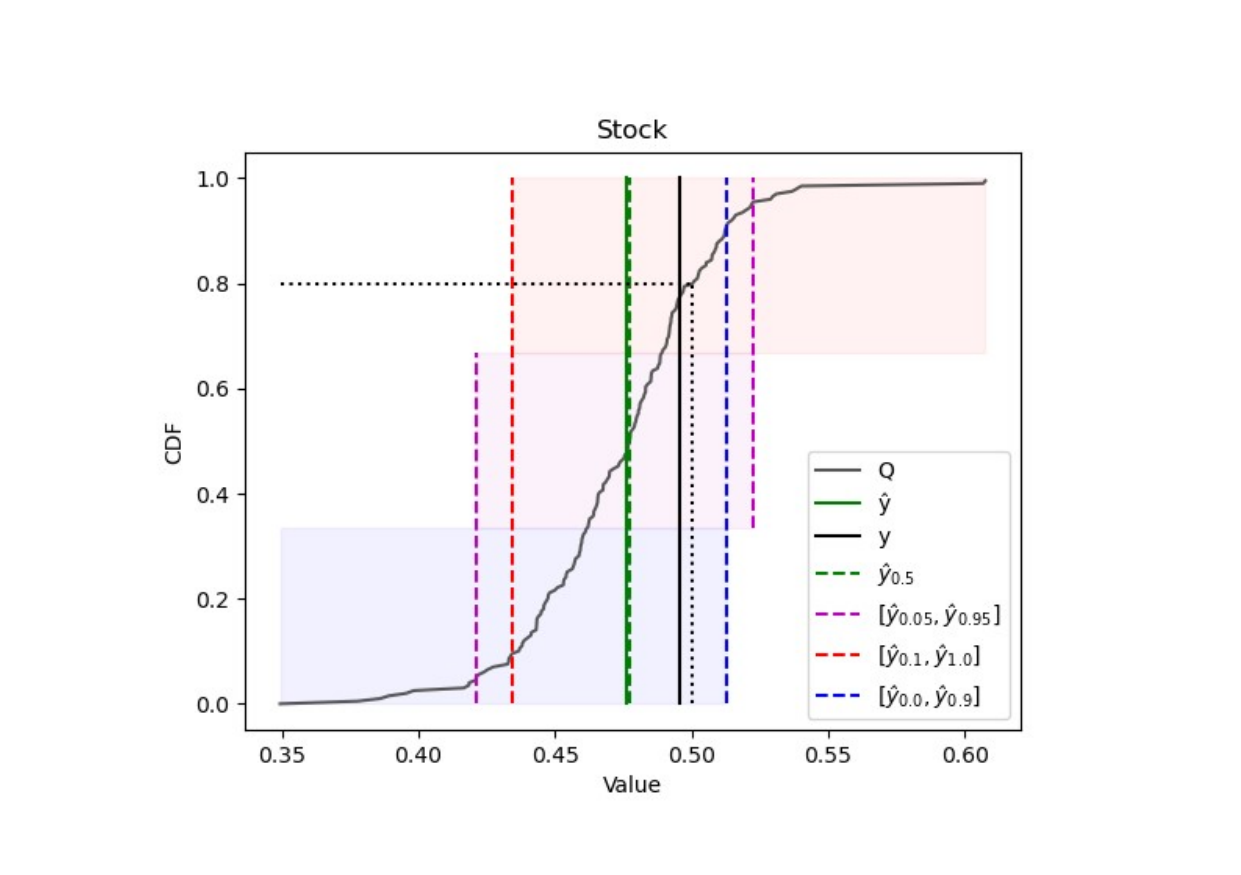}
    \caption{A CPD with three different intervals representing $90\%$ confidence are defined: {\color{red}\textbf{Lower-bounded interval}}: more than the $10^{th}$ percentile; {\color{mypink2}\textbf{Two-sided interval}}: between the $5^{th}$ and the $95^{th}$ percentiles; {\color{blue}\textbf{Upper-bounded interval}}: less than the $90^{th}$ percentile. The black dotted lines indicate how to determine the probability of the true target being smaller than 0.5, which in this case would be approximately $80\%$.}
    \label{fig:csp_explained}
\end{figure}

Figure \ref{fig:csp_explained} illustrates how the CPD can form one-sided and two-sided confidence intervals. It also illustrates how the probability of the true target falling below a given threshold can be determined, as well as connecting a probability with the
threshold it corresponds to.

Compared to a CR, also able to provide valid confidence intervals from the underlying model, a CPS offers richer opportunities to define intervals and probabilities through querying the CPD. One particular strength of a CPS is its ability to calibrate the underlying model. As an example, if the underlying model is consistently overly optimistic, the median from the CPS will adjust for that and provide a calibrated prediction better adjusted to reality.  

There are several different ways that difficulty ($\sigma$) can be estimated, such as:

\begin{itemize}
    \item The (Euclidean) distances to the $k$ nearest neighbors.
    \item The standard deviation of the targets of the $k$ nearest neighbors.
    \item The absolute errors of the $k$ nearest neighbors.
    \item The variance of the predictions of the constituent models, in case the underlying model is an ensemble.
\end{itemize}

\subsection{Factual and Counterfactual Explanations for Regression} 
\label{sec:CER}
\noindent In order to get \ri{factual Calibrated Explanations for regression}, the probability intervals $[\mathcal{P}_l, \mathcal{P}_h]$ and a calibrated probability estimate $\mathcal{P}$ from VA are exchanged for a confidence interval and the median which are derived from the CPD. The confidence interval is defined by user-selected lower and upper percentiles and allows dynamic selection of arbitrary confidence intervals.

Thus, for the algorithm to produce factual and counterfactual rules in the same way as for classification, the only thing that needs to be adjusted in the algorithm described in section~\ref{CEC} is to exchange the calibrator from VA to CPS. Since the confidence interval from CPS is based on the user-provided percentiles, the lower and upper percentiles are two necessary additional parameters. By default, the lower and upper percentiles are $[5^{th}, 95^{th}]$, resulting in a two-sided $90\%$ confidence interval derived from the CPD. One-sided intervals can in practice be handled as a two-sided interval with either $-\infty$ or $\infty$ assigned as lower or upper percentiles. The calibrated probability estimate used in classification is exchanged for the median from the CPD, which in practice represents a calibration of the underlying model's prediction, neutralizing any systematic bias in the underlying model. Consequently, using a CPS effectively enables \ri{factual Calibrated Explanations for regression} with uncertainty quantification of both the prediction from the underlying model and each feature rule.

More formally, the confidence interval and the median are derived as follows:
\begin{enumerate}
    \item Use the calibration set to calculate the calibration residuals $$r_i=y_i - h(x_i), i \in \{1,\dots,q\}.$$
    \item Fit a CPS model $cps$ using the residuals.
    \item Obtain the median and interval values 
    $$\left[m, l, h\right] = cps(h(x), {percentiles}=[50^{th}, \mathcal{P}_l^{th}, \mathcal{P}_h^{th}])$$ 
    using the $50^{th}$, the lower $\mathcal{P}_l^{th}$ and the higher $\mathcal{P}_h^{th}$ percentiles.
    \item To create \ri{factual Calibrated Explanations for regression} following the procedure described in section~\ref{CEC} above, substitute $\mathcal{P}$ and $[\mathcal{P}_l,\mathcal{P}_h]$ from the VA calibrator with $m$ and $[l,h]$ from the CPS.
\end{enumerate}

The input to the
\ri{Calibrated Explanations differs between classification and regression:
in classification, it is probability estimates;
in regression, it is actual predicted values.
Thus, factual Calibrated Explanations for regression}
will result in feature weights indicating changes in predictions rather than changes in probabilities.

\subsection{Factual and Counterfactual Probabilistic Calibrated Explanations for Regression}
\label{sec:PCER}
\noindent The simplest approach when trying to predict the probability that a target value is below (or above) a threshold is to treat the problem as a binary classification problem, with the target defined as 
\begin{equation}
\label{eq:y^c}
    \Dot{y}_i=
    \begin{cases}
        \textstyle 1 & \textrm{if } y_i\leq t\\
        \textstyle 0 & \textrm{if } y_i>t,
    \end{cases}    
\end{equation} where $y$ are the regression targets, $t$ the threshold, and $\Dot{y}$ the binary classification target. To obtain the probability, some form of probabilistic classifier is used. 

The CPS makes it possible to query any regular regression model for the probability of the target falling below any given threshold. This effectively eliminates the need to treat the problem as a classification problem. 

Utilizing this strength to create explanations is straightforward \ri{when only} the probability is of interest. However, there is no obvious equivalent to the probability interval created by VA in classification or the confidence interval derived from a CPS in regression. Consequently, achieving a calibrated explanation with uncertainty quantification for this scenario is not as easy as creating factual and counterfactual explanations for classification or regression. 

The fact that probabilistic predictions for regression can be achieved by viewing it as a classification problem holds a key to a solution. VA needs a score $s$ for both the calibration and the test instances. By using a CPS as a probabilistic scoring function for both calibration and test instances, it becomes possible to use VA to calibrate the probability and provide a probability interval. The score used is the probability (from a CPD) of calibration and test instances being below the given threshold. The isotonic regressors used by VA also need a binary target for the calibration set, which is defined using equation~(\ref{eq:y^c}).

Since the CPS is defined using the calibration set, the probabilities achieved on the same calibration set will be biased and consequently not be entirely trustworthy. To counteract that, \ri{the original calibration set is split in halves, and one half is used as a calibration set for CPS while the other half is used as calibration set for VA. The CPS can be pre-fitted at initialization of the \texttt{CalibratedExplainer} whereas VA needs to be initialized for each threshold at explanation time. More formally, the following is done at initialization: }
\ri{\begin{enumerate}
    \item Split the calibration set into two equal halves and calculate the residuals \\$R=\{r_1,\dots,r_i,\dots,r_{\lceil q/2\rceil}\}$ for the first half, where $r_i=y_i - h(x_i)$.
    \item Fit a CPS model $cps_{\mathcal{P}}$ using the residuals $R$.
\end{enumerate}}

\ri{At explanation time, the following is done:}
\begin{enumerate}
    \item \ri{Define the scores $S=cps_{\mathcal{P}} (\{x_{\lfloor q/2\rfloor+1},\dots,x_q\}, {threshold}=t)$ for the second half of the calibration instances.
    \item Define the categorical targets as $\Dot{Y} = \{y_i\leq t\}$, where $i\in\{\lfloor q/2\rfloor+1,\dots,q\}$.
    \item Use $S$ as scores and $\Dot{Y}$ as targets to define a VA calibrator $va_{\mathcal{P}}$.
    \item Define the score for the test instance $x$ as $s=cps_{\mathcal{P}} (x, {threshold}=t)$.
    \item Use $va_{\mathcal{P}}$ with $s$ to produce} probability intervals $[\mathcal{P}_l, \mathcal{P}_h]$ and a calibrated probability estimate $\mathcal{P}$ for \ri{the true target being below the threshold,} $y\leq t$, and create a calibrated explanation using the description in section~\ref{CEC} \ri{with the class labels $\{y\leq t, y>t\}$}. 
\end{enumerate}
\ri{If the same threshold $t$ is used for a batch of test objects, the same calibrator, $va_{\mathcal{P}}$, is re-used, improving computational performance as the first three steps only needs to be done once. }




\subsection{Properties of Calibrated Explanations for Regression}
\noindent The median from a CPD based on the calibration data can be seen as a form of calibration of the underlying model's prediction, since it may adjust the prediction on the test instance to match what has previously been seen on the calibration set. The calibration will primarily affect systematic bias in the underlying model. Consequently, since \ri{Calibrated Explanations} calibrates the underlying model, it will create calibrated predictions and explanations. In addition, VA provides uncertainty quantification of both the probability estimates from the underlying model and the feature importance weights through the intervals for \ri{probabilistic Calibrated Explanations for regression}. By using equality rules for categorical features and binary rules for numerical features (as recommended above), interpreting the meaning of a rule with a corresponding feature weight in relation to the target and instance value is straightforward and unambiguous and follows the same logic as for classification. 

The explanations are reliable because the rules straightforwardly define the relationship between the calibrated outcome and the feature weight (for \ri{factual explanations}) or feature prediction estimate (for \ri{counterfactual explanations}). The explanations are robust, i.e., consistent, as long as the feature rules cover any perturbations in feature values. Variation in predictions, e.g. when training using different training sets, can be expected to result in some variation in feature rules, corresponding to the variation in predictions. Obviously, the method does not guarantee robustness for perturbations violating a feature rule condition. The \ri{factual and counterfactual Calibrated Explanations for regression} explanations are stable as long as the same calibration set and model are used. Finally, depending on the size of the calibration set which is used to define a CPS, the generation of \ri{factual Calibrated Explanations for regression} is comparable to existing solutions such as LIME and SHAP. Generating a \ri{probabilistic factual Calibrated Explanations for regression} will be slower than \ri{Calibrated Explanations for classification} since both require a VA to be trained. Compared to \ri{Calibrated Explanations for classification}, \ri{probabilistic explanations for regression} will have some additional overhead from using a CPS as well. 

A minor difference between classification and regression is related to the discretizers \ri{that are used for numerical features. Both} the \texttt{BinaryEntropyDiscretizer} and the \texttt{EntropyDiscretizer} (used for classification) require categorical target values for the calibration set \ri{as they use a classification tree (with a depth of $one$ and $three$ levels, respectively) to determine the best discretization. For regression, two new discretizers have been added, \texttt{BinaryRegressorDiscretizer} and \texttt{RegressorDiscretizer}, relying on regression trees, also with depths $one$ and $three$}. The discretizers are automatically assigned based on the kind of problem and explanation that is extracted. The same \texttt{discretizer}s as used for \ri{standard factual and counterfactual Calibrated Explanations for regression} must also be applied for \ri{probabilistic regression explanations}, as it is motivated by the problem type. 

If a difficulty estimator is used to get explanations based on normalized CPDs, $\sigma$ is calculated using \ri{the} \texttt{DifficultyEstimator} in \texttt{crepes.extras} and passed along to $cps$ (and \ri{$cps_{\mathcal{P}}$ for probabilistic regression explanations}) both when fitting and obtaining median and interval values. 

Finally, the calibrated predictions and their confidence intervals, which are an integral part of factual \ri{Calibrated Explanations}, provide the same guarantees as the calibration model used, i.e., the same guarantees as VA for classification and CPSs for regression (or a combination of both for probabilistic regression). However, even if the uncertainty quantification in the form of intervals for the feature rules is also derived from the same calibration model, these feature rule intervals do not necessarily provide the same guarantees. The reason is that the perturbed instances (see \ri{steps~\ref{step3} and~\ref{step4}}) are artificial and the combination of feature values may not always exist naturally in the problem domain. Whenever that happens, the underlying model and the calibration model will indicate that it is a strange instance but may not estimate the degree of strangeness correctly as there is no evidence in the data to base a correct estimate on.

A Python implementation of the \ri{Calibrated Explanations} solution described in this paper is freely available with a BSD3-style license from:

\begin{itemize}[noitemsep,topsep=0pt]
\item Code repository: \href{https://github.com/Moffran/calibrated_explanations}{github.com/Moffran/calibrated\_explanations}
\item PyPi package: \href{https://pypi.org/project/calibrated-explanations/}{pypi.org/project/calibrated-explanations/}
\item Conda-forge package: \href{https://anaconda.org/conda-forge/calibrated-explanations}{anaconda.org/conda-forge/calibrated-explanations}
\item Documentation: \href{https://calibrated-explanations.readthedocs.io/}{calibrated-explanations.readthedocs.io/}
\end{itemize}

Since it is on PyPI and conda-forge, it can be installed with \texttt{pip} or \texttt{conda} commands. The GitHub repository includes Python scripts to run the examples in this paper, making the results here easily replicable. The repository also includes several notebooks with additional examples. This paper details \texttt{calibrated-explanations} as of version \ri{0.3.3}.

\begin{figure}
    \begin{lstlisting}

from calibrated_explanations import CalibratedExplainer
# Load and pre-process your data
# Divide it into proper training, calibration, and test sets

# Train your model using the proper training set
model.fit(X_proper_training, y_proper_training)

# Initialize the CalibratedExplainer
ce = CalibratedExplainer(model, X_calibration, y_calibration, 
                            mode='regression')

# Create and plot factual standard explanations
factual_explanations = ce.explain_factual(X_test, 
                            low_high_percentiles=(5,95))
factual_explanations.plot_all()
factual_explanations.plot_all(uncertainty=True)

# Create and plot counterfactual standard explanations
counterfactual_explanations = ce.explain_counterfactual(X_test, 
                            low_high_percentiles=(5,95))
counterfactual_explanations.plot_all()

# One-sided explanations are easily created
factual_upper_bounded = ce.explain_factual(X_test, 
                            low_high_percentiles=(-np.inf,90))
counterfactual_lower_bounded = ce.explain_counterfactual(X_test, 
                            low_high_percentiles=(10,np.inf))

# Create and plot factual probabilistic explanations
your_threshold = 1000
factual_explanations = ce.explain_factual(X_test, 
                            threshold=your_threshold)

# Create and plot counterfactual probabilistic explanations
counterfactual_explanations = ce.explain_counterfactual(X_test, 
                            threshold=your_threshold)
    \end{lstlisting}
    \caption{\ri{Code example on using \textit{calibrated-explanations} for regression.}}
    \label{fig:code_standard}
\end{figure}

\ri{Using Calibrated Explanations with regression is done using almost identical function calls as for classification. An example on how to initialise a \texttt{CalibratedExplainer} and create factual and counterfactual explanations for standard and probabilistic regression from a trained model can be seen in Fig.~\ref{fig:code_standard}.  The parameter \texttt{low\_high\_percentiles=(5,95)} is the default value and can be left out or changed to some other uncertainty interval. In the example, all intervals are defined to 90\% confidence. The difference between standard and probabilistic explanations only require exchanging \texttt{low\_high\_percentiles=(low,high)} with \texttt{threshold=your\_threshold}. The \texttt{threshold} parameter is \texttt{None} by default but takes precedence when having a value assigned.}

\begin{figure}
    \begin{lstlisting}
from calibrated_explanations import CalibratedExplainer
from crepes.extras import DifficultyEstimator 
# Load and pre-process your data
# Divide it into proper training, calibration, and test sets

# Train your model using the proper training set
model.fit(X_proper_training, y_proper_training)

de = DifficultyEstimator()

# 1: by the (Euclidean) distances to the nearest neighbors
de.fit(X=X_proper_training, scaler=True)

# 2: by the standard deviation of the targets of the nearest 
#    neighbors
de.fit(X=X_proper_training, y=y_proper_training, scaler=True)

# 3: by the absolute errors of the k nearest neighbors
residuals_oob = y_proper_training - model.oob_prediction_
de.fit(X=X_proper_training, residuals=residuals_oob, scaler=True)

# 4: by the variance among ensemble submodels
de.fit(X=X_proper_training, learner=model, scaler=True)

# Initialize the CalibratedExplainer with de
ce = CalibratedExplainer(model, X_calibration, y_calibration, mode='regression', difficulty_estimator=de)

# Change a DifficultyEstimator
ce.set_difficultyEstimator(de)
    \end{lstlisting}
    \caption{\ri{Code example on using \textit{calibrated-explanations} with normalization.}}
    \label{fig:code_normalized}
\end{figure}
\ri{Normalization can be achieved using \texttt{DifficultyEstimator} from \texttt{crepes.extras}. It currently has four different ways to normalize, as seen in the example shown in Fig.~\ref{fig:code_normalized}, where alternative 3 and 4 requires an ensemble model, such as a \texttt{RandomForestRegressor}. Creating normalized explanations with standard and probabilistic regression is done exactly the same as without normalization, see Fig.~\ref{fig:code_standard}, once the difficulty estimator is assigned.}

\subsection{Summary of \ri{Calibrated Explanations}}
\noindent With the two solutions proposed here, \ri{Calibrated Explanations} provide a number of possible use cases, which are summarized in Table~\ref{tab:summary}. 

\begin{table}[htpb]
    \centering
    \begin{tabular}{l|ccc|ccc|ccc}
         & \multicolumn{3}{c|}{Classification} & \multicolumn{3}{c|}{Standard} & \multicolumn{3}{c}{Probabilistic} \\
         & \multicolumn{3}{c|}{} & \multicolumn{3}{c|}{Regression} & \multicolumn{3}{c}{Regression} \\
        Characteristics & FR & FU & CF & FR & FU & CF & FR & FU & CF \\
        \hline
        Feature Weight w/o CI & X &  &  & X &  &  & X &  &  \\
        Feature Weight with CI &  & X &  &  & X &  &  & X &  \\
        Rule Prediction with CI &  &  & X &  &  & X &  &  & X  \\
        \hline
        Two-sided CI & I & I & I & I & I & I & I & I & I  \\
        Lower-bounded CI &  &  &  & I &  & I &  &  &  \\
        Upper-bounded CI &  &  &  & I &  & I &  &  &  \\
        \hline
        \ri{Conjunctive rules} & O & O & O & O & O & O & O & O & O  \\
        \ri{Conditional rules} & O & O & O & O & O & O & O & O & O  \\
        Normalization &  &  &  & O & O & O & O & O & O  \\
        \# alternative setups & 1 & 1 & 1 & 5 & 5 & 5 & 5 & 5 & 5  \\ 
    \end{tabular}
    \caption{Summary of characteristics of \ri{Calibrated Explanations. All explanations include the \textit{calibrated prediction}, with \textit{confidence intervals}, of the explained instance. \textbf{FR} refers to \textit{factual} explanations visualized using \textit{regular} plots, \textbf{FU} refers to \textit{factual} explanations visualized using \textit{uncertainty} plots, and \textbf{CF} refers to \textit{counterfactual} explanations and plots. Furthermore, \textbf{CI} refers to a \textit{confidence interval}, \textbf{Conjunctive rules} indicates that \textit{conjunctive rules are possible}, \textbf{Conditional rules} indicates \textit{support for users to create contextual explanations}, \textbf{Normalization} indicates that \textit{normalization is supported} and \textbf{\#~alternative setups} refers to \textit{the number of ways to run Calibrated Explanations}, i.e., w/o normalization or with any of the four different ways to normalize. \textbf{X} marks a \textit{core alternative}, \textbf{I} marks \textit{selectable interval type(s)} used by the core alternatives, and \textbf{O} marks \textit{optional additions}.}}
    \label{tab:summary}
\end{table}

Both factual and counterfactual explanations are composed of lists of feature rules with conditions and feature weights with confidence intervals (factual explanations) or feature prediction estimates with confidence intervals (counterfactual explanations), as described in Section~\ref{CE}. \ri{Conditional rules was introduced in version 0.3.1 and described in a paper introducing this for analysis of Fairness \cite{lofstrom2024fairness}}.

\section{Experimental Setup}
\noindent The implementation of both the regression and the probabilistic regression solutions is expanding the \texttt{calibrated-explanations} Python package \citep{calibrated-explanations} and relies on the \texttt{ConformalPredictiveSystem} from the \texttt{crepes} package \citep{crepes}. By default, \texttt{ConformalPredictiveSystem} is used without normalization but \texttt{DifficultyEstimator} provided by \texttt{crepes.extras} is fully supported by \texttt{calibrated-explanations}, with normalization options corresponding to the list given at the end of Section~\ref{sec:cps} \ri{and in Fig.~\ref{fig:code_normalized}}. 

\subsection{Presentation of Calibrated Explanations trough Plots}
\noindent In this paper, three different kinds of plots for \ri{Calibrated Explanations} are presented. The first two are used when visualizing \ri{factual Calibrated Explanations for standard regression}. These plots are inspired by LIME, especially the rules in LIME have been seen as providing valuable information in the explanations. 

\begin{itemize}
    \item Regular explanations, providing \ri{Calibrated Explanations} without any uncertainty information. These explanations are directly comparable to other feature importance explanation techniques like LIME. 
    \item Uncertainty explanations, providing \ri{Calibrated Explanations} including uncertainty intervals to highlight both the importance of a feature and the amount of uncertainty connected with its estimated importance.
\end{itemize} 

For the reasons given in previous sections, \ri{Calibrated Explanations} is meant to use binary rules with factual explanations (even if all discretizers used by LIME can also be used by \ri{Calibrated Explanations}). One noteworthy aspect of \ri{Calibrated Explanations} is that the feature weights only show how each feature separately affects the outcome. It is possible to see combined weights through conjunctions of features (combining two or three different rules into a conjunctive feature rule). It is important to clarify that the feature weights do not convey the same meaning as in attribution-based explanations, like SHAP.

The third kind of plot is a counterfactual plot showing preliminary prediction estimates for each feature when alternative feature values are used.

Feature rules are always ordered based on feature weight, starting with the most impactful rules.  When plotting \ri{Calibrated Explanations}, the user can choose to limit the number of rules to show. Factual explanations have one rule per feature. Counterfactual explanations, where \ri{Calibrated Explanations} creates as many counterfactual rules as possible, may result in a much larger number of rules, especially for categorical \ri{features} with many categories.

Internally, \ri{Calibrated Explanations} uses the same representation for both classification and regression. However, the plots visualizing the explanations have been adapted to suit \ri{both standard and probabilistic factual Calibrated Explanations for regression}. 

\subsubsection{Calibrated Explanations Plots}
\noindent The same kind of plots exists for regression as for classification. Compared to the plots used for classification, the regression plots differ in two essential aspects.

A common difference for both \ri{factual and counterfactual Calibrated Explanations for regression} is that the feature weights represent changes in actual target values. For \ri{factual Calibrated Explanations for regression}, this means that a feature importance of $+100$ means that the actual feature value contributes with $+100$ to the prediction. For a \ri{counterfactual Calibrated Explanations for regression}, showing the prediction estimates with uncertainty intervals, the plot shows what the prediction is estimated to have been if the counterfactual condition would be fulfilled. 

A difference that only applies to the factual plots is that the top of the plot omits the probabilities for the different classes and instead shows the median $m$ and the confidence interval $[l,h]$ as the prediction.

\subsubsection{Probabilistic Calibrated Explanations Plots}
\noindent Since the \ri{probabilistic factual Calibrated Explanations for regression} represents feature importances as probabilities, just like \ri{Calibrated Explanations for classification}. The only difference needed for the \ri{probabilistic} plots for \ri{regression} compared to classification is to change the probabilities for a class label into probabilities for being below ($\mathcal{P}(y \leq t)$) or above ($\mathcal{P}(y > t)$) the given threshold.

\subsection{Experiments} 
\noindent The evaluation is divided into an introduction to \ri{all different kinds of Calibrated Explanations for regression} through plots and an evaluation of performance. All plots are from the California Housing data set \citep{california-housing}. The underlying model in all experiments is a \texttt{RandomForestRegressor} from the \texttt{sklearn} package.

Our proposed algorithm is claimed to be fast, reliable, stable, and robust. These claims requires validation in an evaluation of performance. The explanations are reliable due to the validity of the uncertainty estimates used, i.e., the results achieved by querying the CPD, and from the uncertainty quantification of the feature weights or feature prediction estimates. Speed, stability and robustness will be evaluated in an experiment using the California Housing data set on a fixed set of test instances. Each experiment is repeated $100$ times using $500$ instances as a calibration set \ri{(also used by SHAP and LIME)} and $10$ test instances. The target values were normalized, i.e., $y\in [0,1]$. The following setups are evaluated:
\begin{itemize}
    \item FCER: Factual explanation. 
    \item CCER: Counterfactual explanation.
    \item PFCER: Probabilistic factual explanation. The threshold is $0.5$ for all instances, i.e., the mid-point of the interval of possible target values.    
    \item \ri{PCCER: Probabilistic counterfactual explanation. The threshold is $0.5$ for all instances, i.e., the mid-point of the interval of possible target values.}
    \item LIME: LIME explanation.
    \item LIME CPS: LIME explanation using the median from a CPD as prediction. The CPS was based on the underlying random forest regressor.
    \item SHAP: SHAP explanation using the \texttt{Explainer} class.
    \item SHAP CPS: SHAP explanation using the median from a CPD as prediction. The CPS was based on the underlying random forest regressor. Here, the \texttt{Explainer} class was used.
\end{itemize}

\noindent The evaluated metrics are:
\begin{itemize}    
    \item \textit{Stability} means that multiple runs on the same instance and model should produce consistent results. Stability is evaluated by generating explanations for the same predicted instances 100 times with different random seeds \ri{(using the iteration counter as random seed). The random seed is used to initialize the \texttt{numpy.random.seed()} and by the discretizers}. 
    The largest variance in feature weight (or feature prediction estimate) can be expected among the most important features (by definition \ri{of} having higher absolute weights). 
    The top feature for each test instance is identified as the feature being most important most often in the 100 runs (i.e., the mode of the feature ranks defined by the absolute feature weight). The variance for the top feature is measured over the 100 runs and the mean variance among the test instances is reported. 

    \item \textit{Robustness} means that small variations in the input should not result in large variations in the explanations. Robustness is measured in a similar way as stability, but with the training and calibration set being randomly drawn and a new model being fitted for each run, creating a natural variation in the predictions of the same instances without having to construct artificial instances. Again, the variance of the top feature is used to measure robustness. The same setups as for stability are used except that each run use a new model and calibration set and that the random seed was set to $42$ in all experiments.    
    
    \item \ri{\textit{Run time}} is compared between the setups regarding explanation generation times (in seconds per instance). It is only the method call resulting in an explanation that is measured. Any overhead in initiating the explainer class is not considered). The closest equivalent to \ri{probabilistic factual Calibrated Explanations for regression} would be to apply LIME and SHAP for classification to a thresholded classification model, as described in section~\ref{sec:PCER}. Since VA is comparably slow and \ri{probabilistic Calibrated Explanations for regression} combines both CPSs and VA, with fitting and calls to a CPS for each calibration instance, it can be expected to be slow. 
\end{itemize}

\ri{FCER and PFCER without normalization are compared with the LIME and SHAP alternatives. Additionally, \ri{run time} is compared across both standard and probabilistic factual and counterfactual Calibrated Explanations with and without normalization. The difficulty estimation uses $500$ randomly drawn instances from the training set to estimate difficulty. Stability and robustness are less affected by normalization\footnote{Detailed results for stability and robustness can be found in the \href{https://github.com/Moffran/calibrated_explanations/tree/main/evaluation/regression}{\textit{evaluation/regression}} folder in the repository, together with the code used for experiments shown in the paper.}. }

\section{Results} 
\noindent The results are divided into two parts: 1) a presentation of \ri{Calibrated Explanations} through plots, explaining and showcasing a number of different available ways \ri{Calibrated Explanations} can be used and viewed; and 2) an evaluation of performance with comparisons to LIME and SHAP.

\subsection{Presentation of Calibrated Explanations through Plots}
\noindent In the following subsections, a number of introductory examples of \ri{Calibrated Explanations} are given for regression. First, factual and counterfactual explanations for regression are shown, followed by factual and counterfactual explanations for probabilistic regression. 

\subsubsection{Factual Calibrated Explanations for Regression}
\begin{figure}[htbp]
    \centering
    \includegraphics[width=0.9\textwidth, trim={0 0.9cm 0 0.2cm},clip]{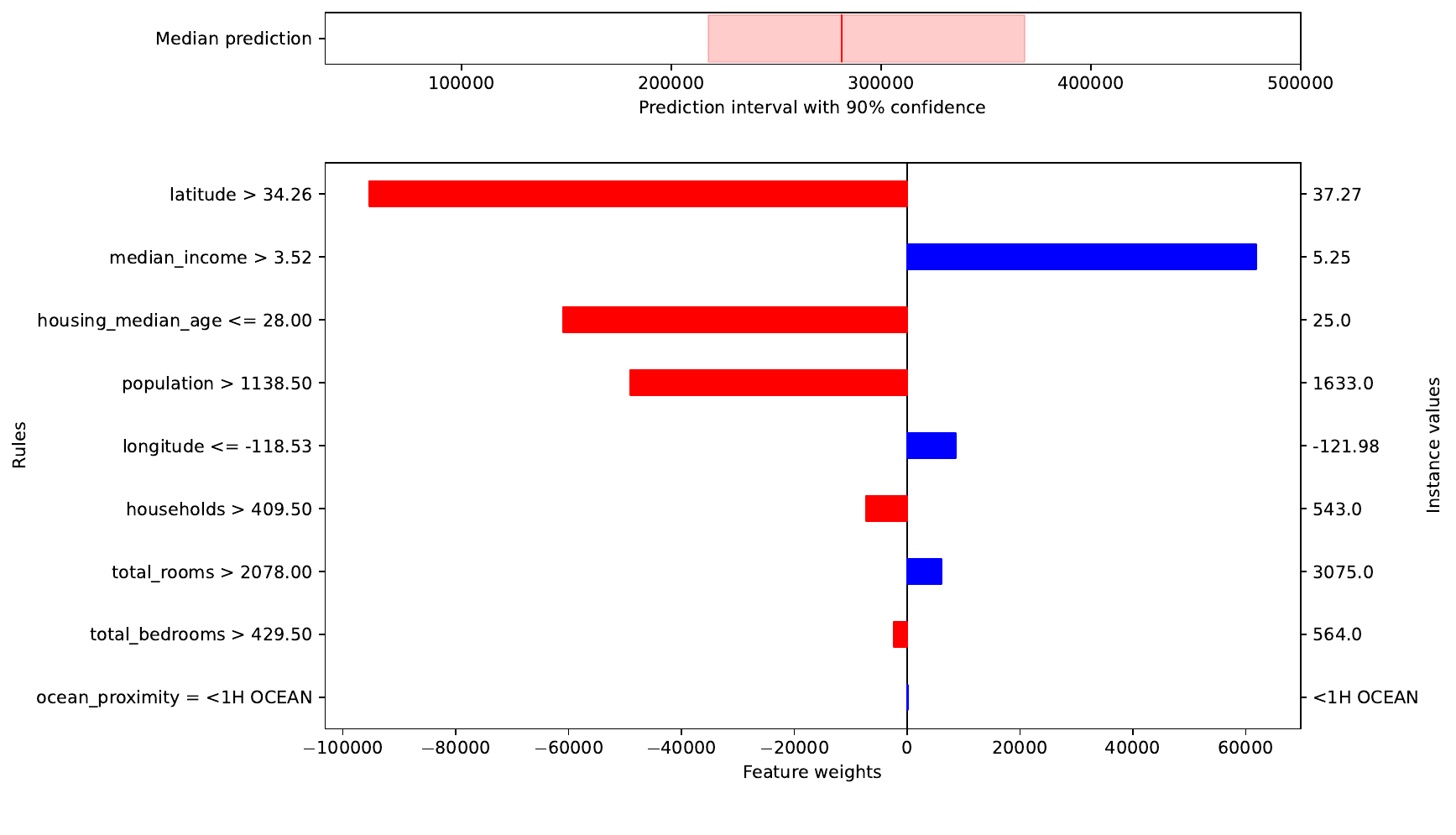}
    \caption{A regular plot for the California Housing data set. The top-bar illustrates the median (the red line) and a confidence interval (the light red area), defined by the $5^{th}$ and the $95^{th}$ percentiles. The subplot below visualizes the weights associated with each feature. The weights indicate how much that rule contributes to the prediction. Negative weights \ri{in red} indicate a negative impact on the prediction whereas positive weights \ri{in blue} indicate a positive impact.}
    \label{fig:housing_simple}
\end{figure}

\noindent The regular plot in Fig.~\ref{fig:housing_simple} illustrates the calibrated prediction of the underlying model as the solid red line at the top bar together with the $90\%$ confidence interval in light red. As can be seen, the house price is predicted to be $\approx$ \$285K and with $90\%$ confidence, the price can be expected to be between [\$215K-\$370K]. Turning to the feature rules, the solid black line represents the median in the top-bar. The rule condition is shown to the left and the actual instance value is shown to the right of the lower plot area. The fact that this house is located more northbound (\texttt{latitude > 34.26}) has a large negative impact on the price (reducing it with $\approx$ \$95K). On the other hand, since the \textit{median income} is a bit higher (\texttt{median income > 3.52}), the price is pressed about \$60K upwards. \texttt{Housing median age} and \texttt{population} are two more features that clearly impact the price negatively. 
\label{sec:CER-res}

\begin{figure}[htbp]
  \centering
  \subfloat[Upper bounded explanation]{\includegraphics[width=0.8\textwidth,trim={0.3cm 0 0.55cm 0},clip]{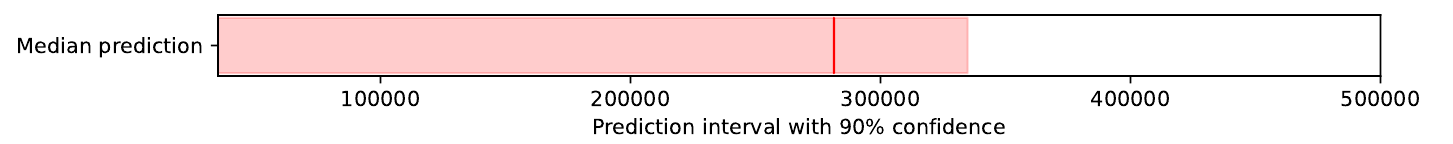}\label{fig:simple_one_1}}
  \vfill
  \subfloat[Lower bounded explanation]{\includegraphics[width=0.8\textwidth,trim={0.2cm 0 0.1cm 0},clip]{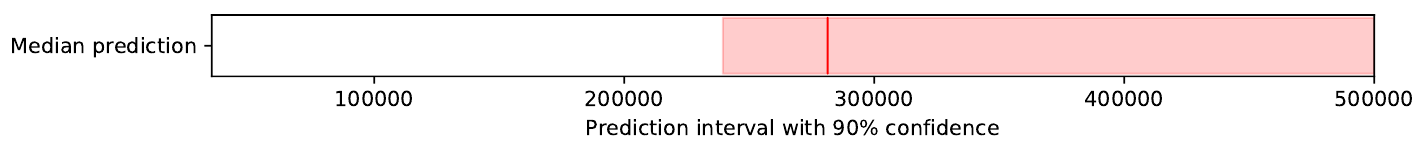}\label{fig:simple_one_2}}
  \caption{The top bars of one-sided plots with confidence intervals bounded by the $90^{th}$ upper percentile (Fig.~\ref{fig:simple_one_1}) and the $10^{th}$ lower percentile (Fig.~\ref{fig:simple_one_2}). The red solid line represents the median. The weights (and consequently the entire subplot visualizing weights) are the same for these one-sided explanations as in Fig.~\ref{fig:housing_simple}.}
\end{figure}

When one-sided intervals are used instead, only the top-bar is affected compared to when using regular plots. Figures~\ref{fig:simple_one_1} and~\ref{fig:simple_one_2} illustrate an upper bounded and a lower bounded explanation for the same instance, with the identical feature rule subplot omitted. As can be seen, the median (solid red line) is the same as before, while the confidence interval stretches one entire side of the bar. The upper bound ($\approx$ \$330K in Fig.~\ref{fig:simple_one_1}) is lower and the lower bound ($\approx$ \$240K in Fig.~\ref{fig:simple_one_2}) is higher compared to the two-sided plot in Fig.~\ref{fig:housing_simple}. 

\label{sec:CUER-res}
\begin{figure}[htbp]
    \centering
    \includegraphics[width=0.9\textwidth, trim={0 0.9cm 0 0.2cm},clip]{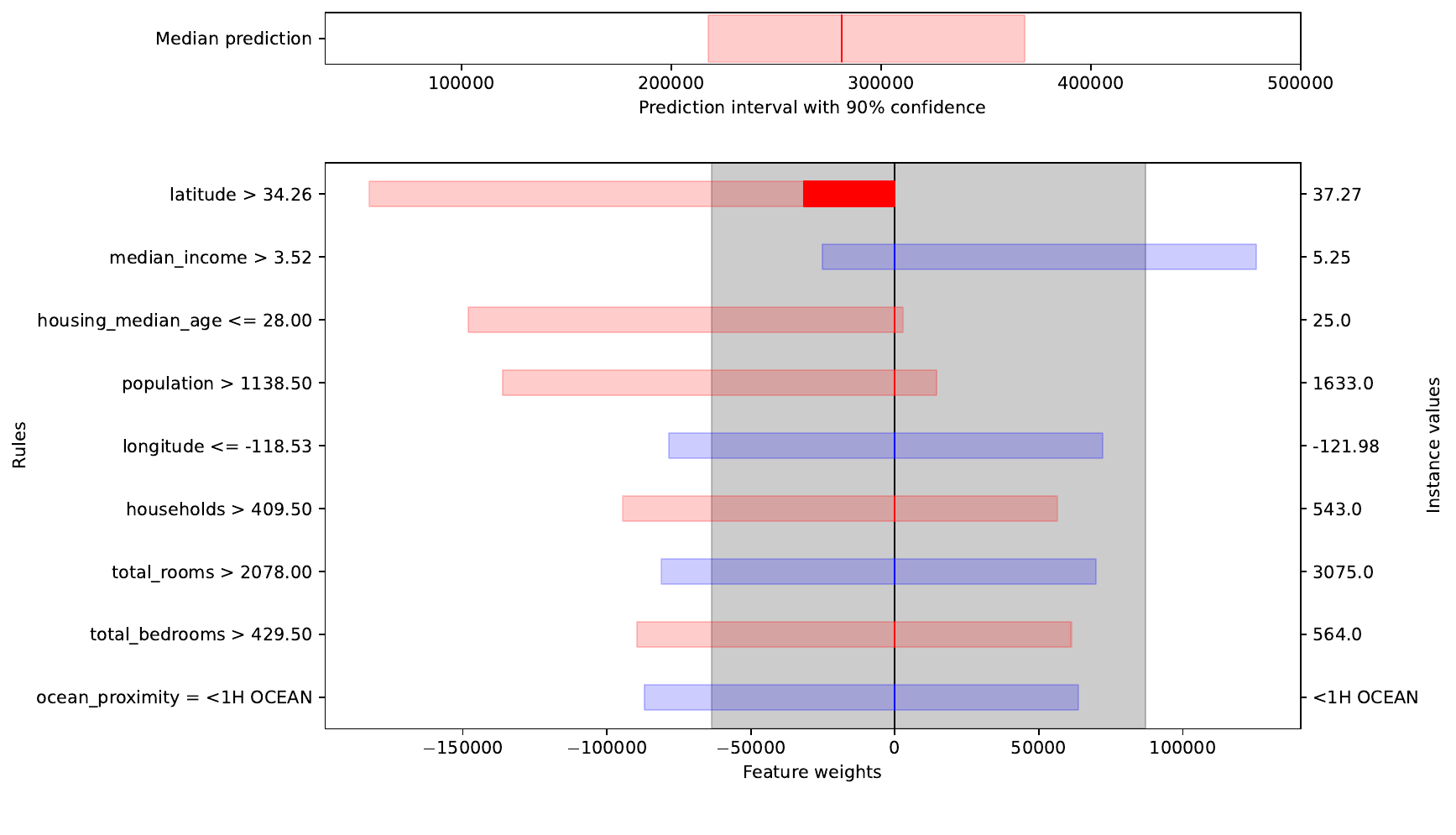}
    \caption{An uncertainty plot for the California Housing data set. The top bar is the same as in Fig.~\ref{fig:housing_simple}, showing the median and the $[5^{th},95^{th}]$ percentiles confidence interval. In the subplot below, the uncertainty of the weights is highlighted, using the $[5^{th},95^{th}]$ percentiles confidence interval in light red or blue for each feature. The weights still indicate how much that rule contributes to the prediction but with a confidence interval highlighting the span of uncertainty for the impact of the feature value and rule combined.}
    \label{fig:housing_interval}
\end{figure}

Fig.~\ref{fig:housing_interval} illustrates an uncertainty plot for the same instance as before\footnote{Uncertainty plots are not available for one-sided explanations, as the visualization becomes obscured and hard to interpret. However, the one-sided uncertainty interval for each feature rule is calculated and can be accessed and used if needed.}. When including uncertainty quantification in the plot, the feature importance has a light colored area corresponding to the span of possible contribution within the confidence used. The grey area surrounding the solid black line represents the same confidence interval as seen in the top bar. 

As can be seen, the northbound \textit{location} still has a large negative impact but the span of uncertainty about exactly how large the impact \ri{is} covers about \$150K, falling approximately within the interval [-\$180K, -\$30K]. The fact that part of the line is solid in color indicates that we can expect this feature to impact the price at least with -\$30K, given the selected confidence level. Looking at the other features, we can see that all of them include the median in the uncertainty interval, meaning that with 90\% confidence, these features may impact the price in both directions. Obviously, both \textit{median income} and in particular \textit{housing median age} are more likely to have a positive and negative impact, respectively. Since no normalization have been used with this example, all the intervals are similar in width.   

\label{sec:CCER-res}
\begin{figure}[htbp]
    \centering
    \includegraphics[width=0.9\textwidth]{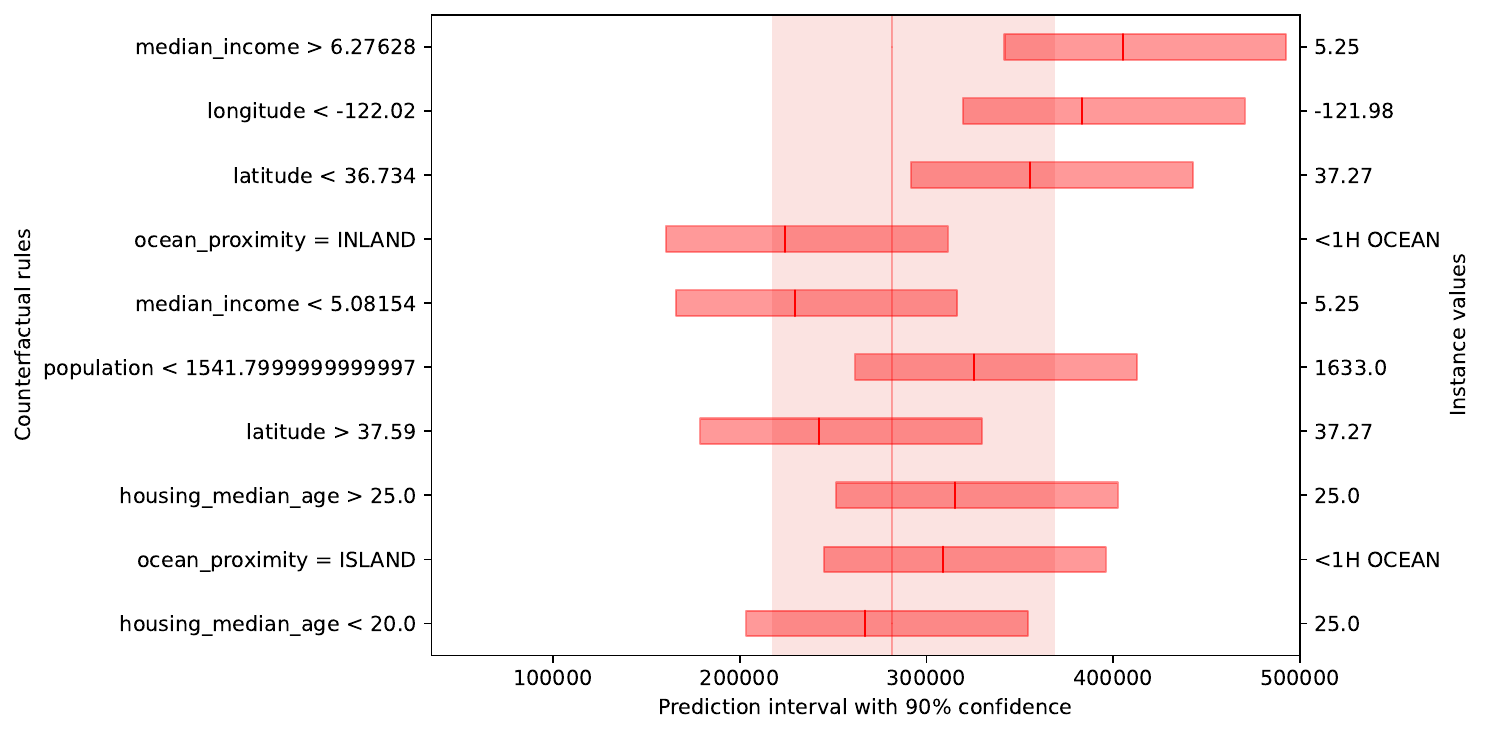}
    \caption{A \ri{counterfactual} plot for the California Housing data set. The large lightest red area in the background is the confidence interval defined by the $5^{th}$ and the $95^{th}$ percentiles. Each row represents a counterfactual rule with an interval in darker red indicating what confidence intervals a breach according to the rule condition would result in. The confidence intervals for the counterfactual rules are also defined by the $5^{th}$ and the $95^{th}$ percentiles. The solid lines represent the median values.}
    \label{fig:housing_simple_counter}
\end{figure}

\subsubsection{Counterfactual Calibrated Explanations for Regression}
\noindent Turning to \ri{counterfactual Calibrated Explanations for regression}, Fig.~\ref{fig:housing_simple_counter} shows a \ri{counterfactual} plot for the same instance as before. Here, the solid line and the very light area behind it represent the median and the confidence interval of the calibrated prediction of the underlying model (i.e., the same as in Fig.~\ref{fig:housing_simple}). This is the ground truth that all the counterfactual feature rules should be contrasted against.

Contrary to \ri{factual Calibrated Explanations for regression}, none of the rules cover the instance values in the \ri{counterfactual} plot. Instead, there are several examples of the same feature being present in multiple rules. Here the interpretation is that the solid line and lighter red bar for each rule is the expected median and confidence interval achieved if the instance would have had values according to the rule. As an example, with everything else the same but \texttt{median income > 6.28}, then the expected price would be $\approx$ \$405K with a confidence interval of [\$340K, \$490K]. It is also clear that if the house would have been located further south (\texttt{latitude < 36.7}), the price would go up, and if it would have been even further north (\texttt{latitude > 37.6}), the price would have gone down even further. \ri{It is worth noting that as the counterfactual rules presented in Fig.~\ref{fig:housing_simple_counter} are excluding the instance values, whereas the factual rules in Figs.~\ref{fig:housing_simple} and~\ref{fig:housing_interval} are including the instance values, the ordering of features may be completely different between the explanations, despite explaining the same instance.}

\begin{figure}[htbp]
    \centering
    \includegraphics[width=0.9\textwidth]{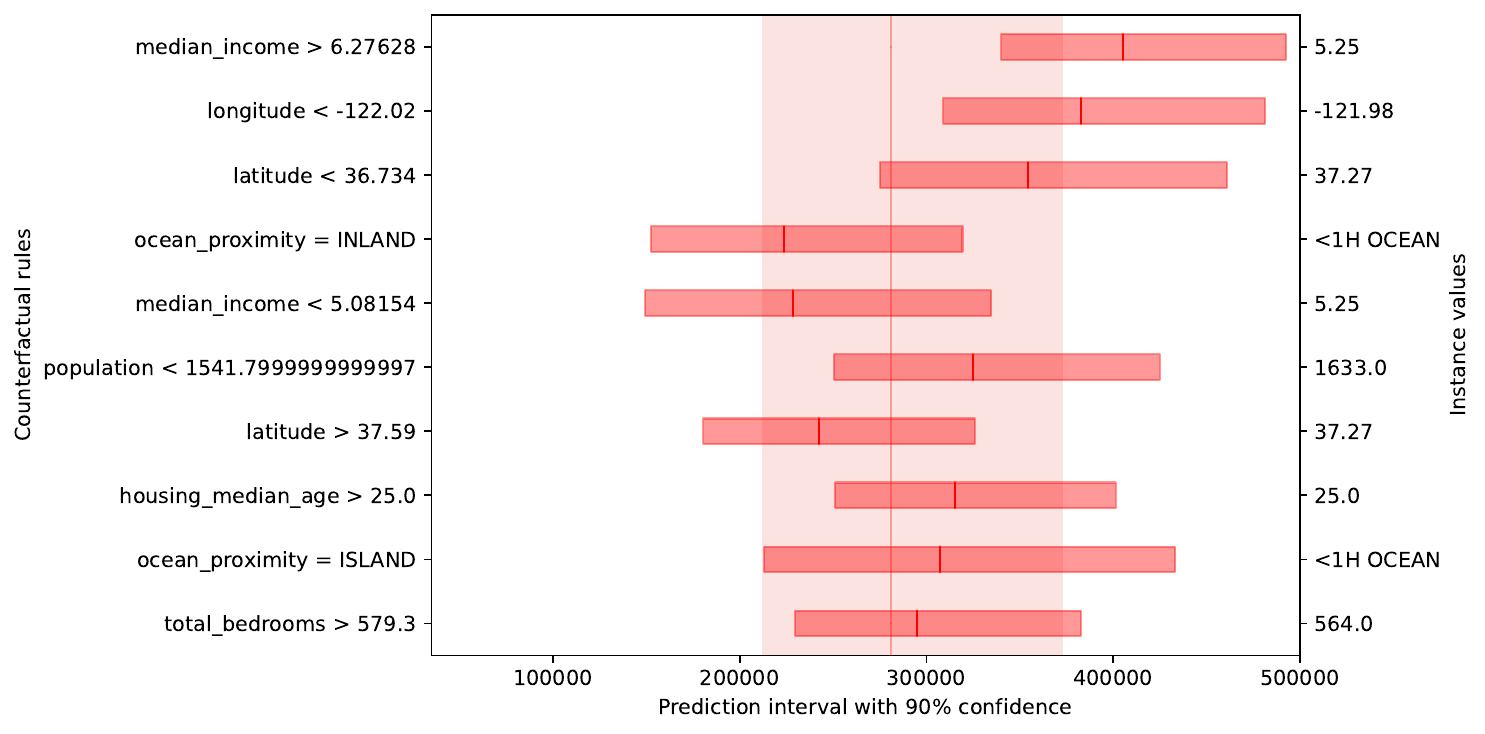}
    \caption{A normalized \ri{counterfactual} plot comparable to Fig~\ref{fig:housing_simple_counter}, resulting in rules with varied interval widths as a consequence of the normalization. Difficulty is estimated as the standard deviation of the targets of the k nearest neighbors.}
    \label{fig:housing_knn_counter}
\end{figure}

So far, all examples \ri{(using both factual and counterfactual explanations)} have used a standard CPS to construct the explanations, with the result that all confidence intervals are almost equal-sized. In Fig.~\ref{fig:housing_knn_counter}, a difficulty estimator based on the standard deviation of the targets of the k nearest neighbors is used. The normalization will both affect the calibration of the underlying model, creating confidence intervals with varying sizes between instances, and the feature intervals. A crude assumption regarding the width of the feature intervals is that when the calibration set contains fewer instances covering an alternative feature value, the feature intervals will tend to be larger due to less information, and vice versa. This does not have to be the whole truth, as difficulty in this example is defined based on the standard deviation of the neighboring instances target values. As can be seen in Fig.~\ref{fig:housing_knn_counter}, normalized counterfactual explanations may generate rules resulting in both smaller and wider confidence intervals then the non-normalized rules. 

\begin{figure}[htbp]
    \centering
    \includegraphics[width=0.9\textwidth]{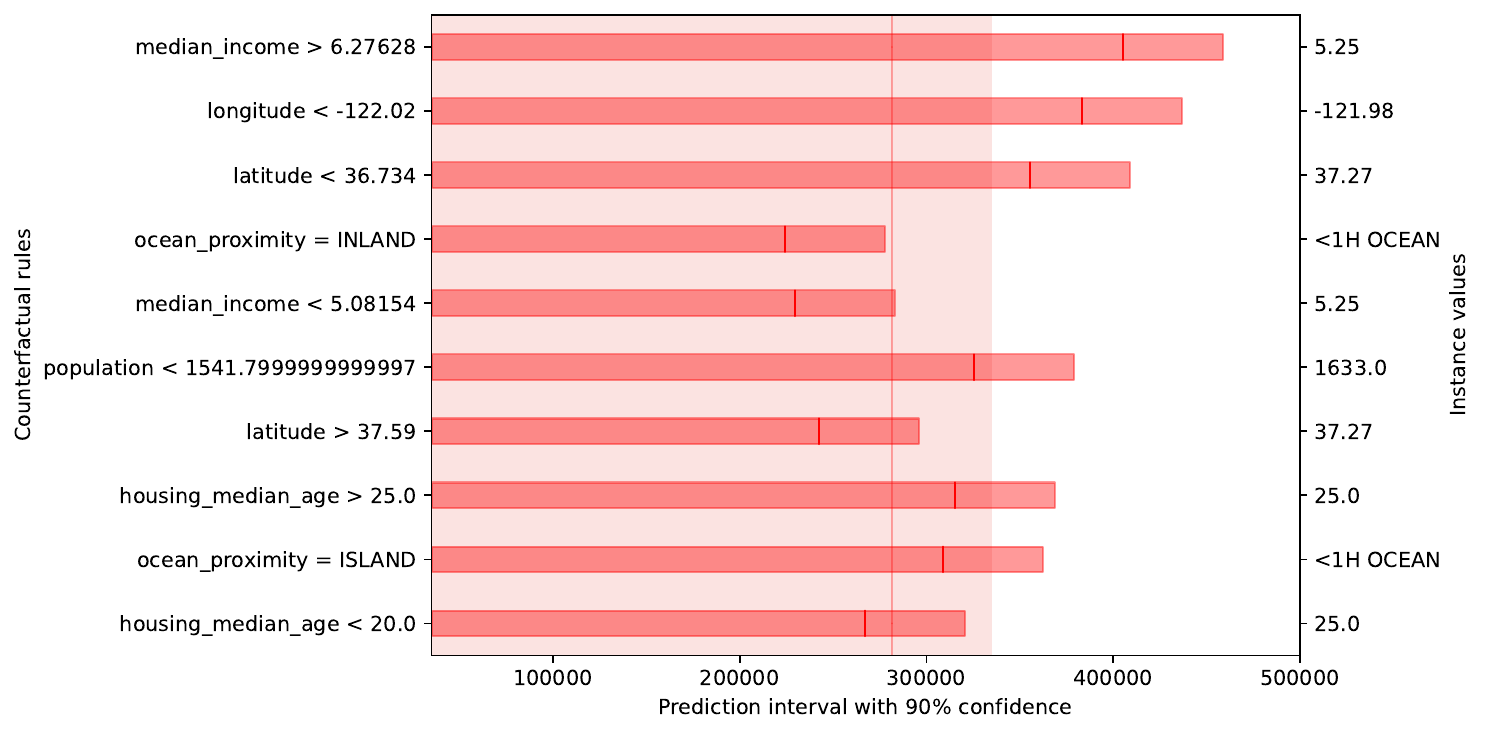}
    \caption{A one-sided \ri{counterfactual} plot for the California Housing data set. Confidence intervals are defined by the $90^{th}$ upper percentile only. The interpretation is that with $90\%$ certainty, the true value of the original instance will fall within the lightest red area. If the counterfactual rule had been true for each feature individually, the true value will fall within that feature's darker red area with approximately $90\%$ certainty.}
    \label{fig:housing_upper_counter}
\end{figure}

Similarly to \ri{factual Calibrated Explanations for regression, counterfactual explanations} can also be one-sided. Fig.~\ref{fig:housing_upper_counter} shows an upper-bounded explanation with $90\%$ confidence. The interpretation of the first rule is that, with everything else as before, but \texttt{median income > 6.28} the price will be below $\approx$ \$450K with 90\% certainty. Since the same CPS is used, the median is still the same as for a two-sided explanation.

\subsubsection{Probabilistic Factual Calibrated Explanations for Regression}
\label{sec:CPER-res}
\begin{figure}[htbp]
    \centering
    \includegraphics[width=0.9\textwidth, trim={0 0.8cm 0 0.2cm},clip]{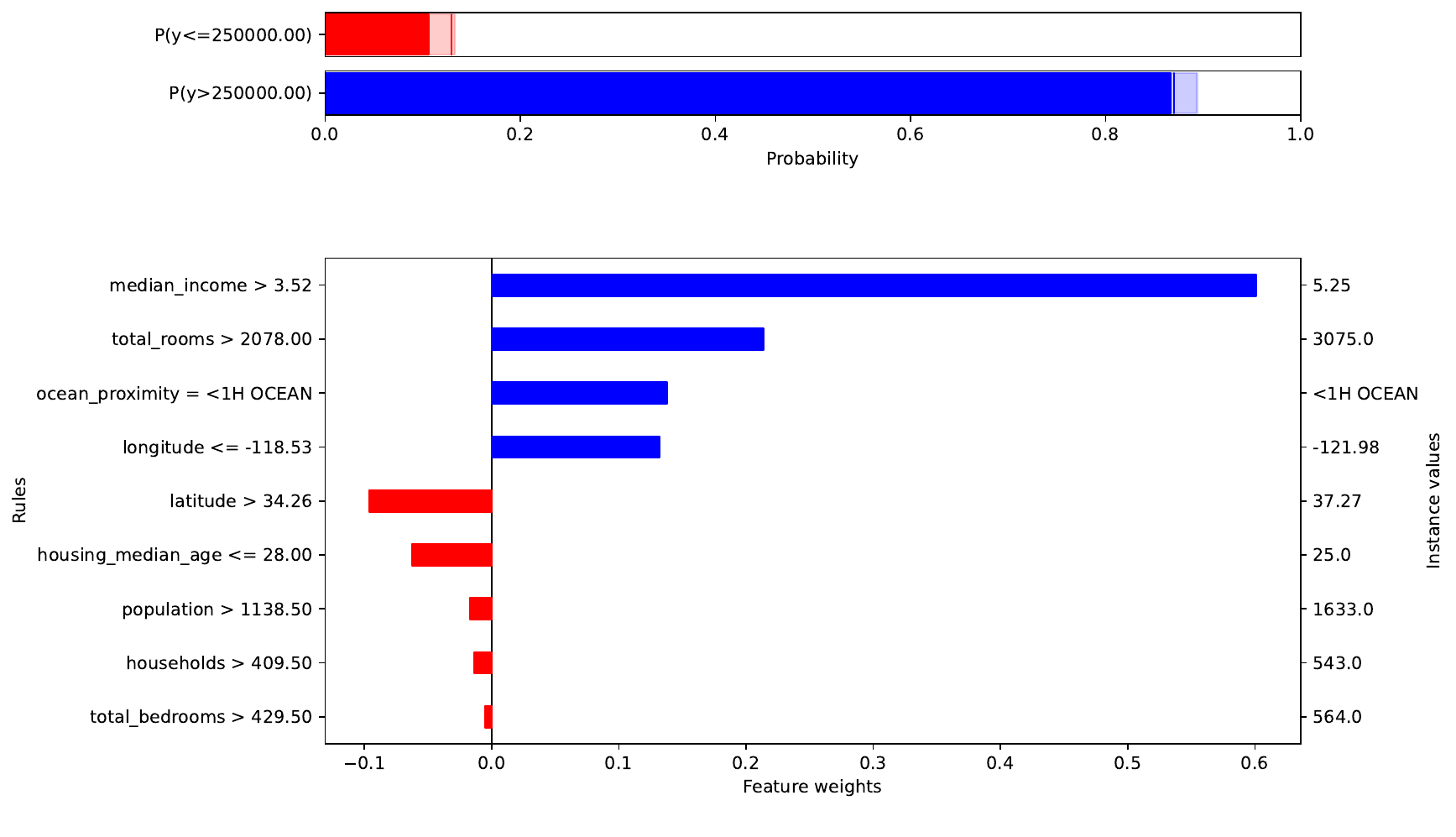}
    \caption{A regular \ri{probabilistic regression} plot for the California Housing data set. The plot shows the probability of the prediction for this instance being above the given threshold (\$250K in this case). The explanation is similar to a regular \ri{plot used in \ri{Calibrated Explanations for classification}} with the main difference being that it shows the probabilities of being below or above the threshold and that the probabilities are given by the CPD.}
    \label{fig:housing_probability} 
\end{figure}
    
Fig.~\ref{fig:housing_probability} shows a regular \ri{probabilistic regression} plot for the same instance as above. In this plot, the possibility of querying the CPD about the probability of being below or above a given threshold is utilized. In this case, the threshold is set to a house price of \$250K. Here, \texttt{median income > 3.52} contributes strongly to the probability that the target is above \$250K.

\begin{figure}[htbp]
    \centering
    \includegraphics[width=0.9\textwidth, trim={0 0.8cm 0 0.2cm},clip]{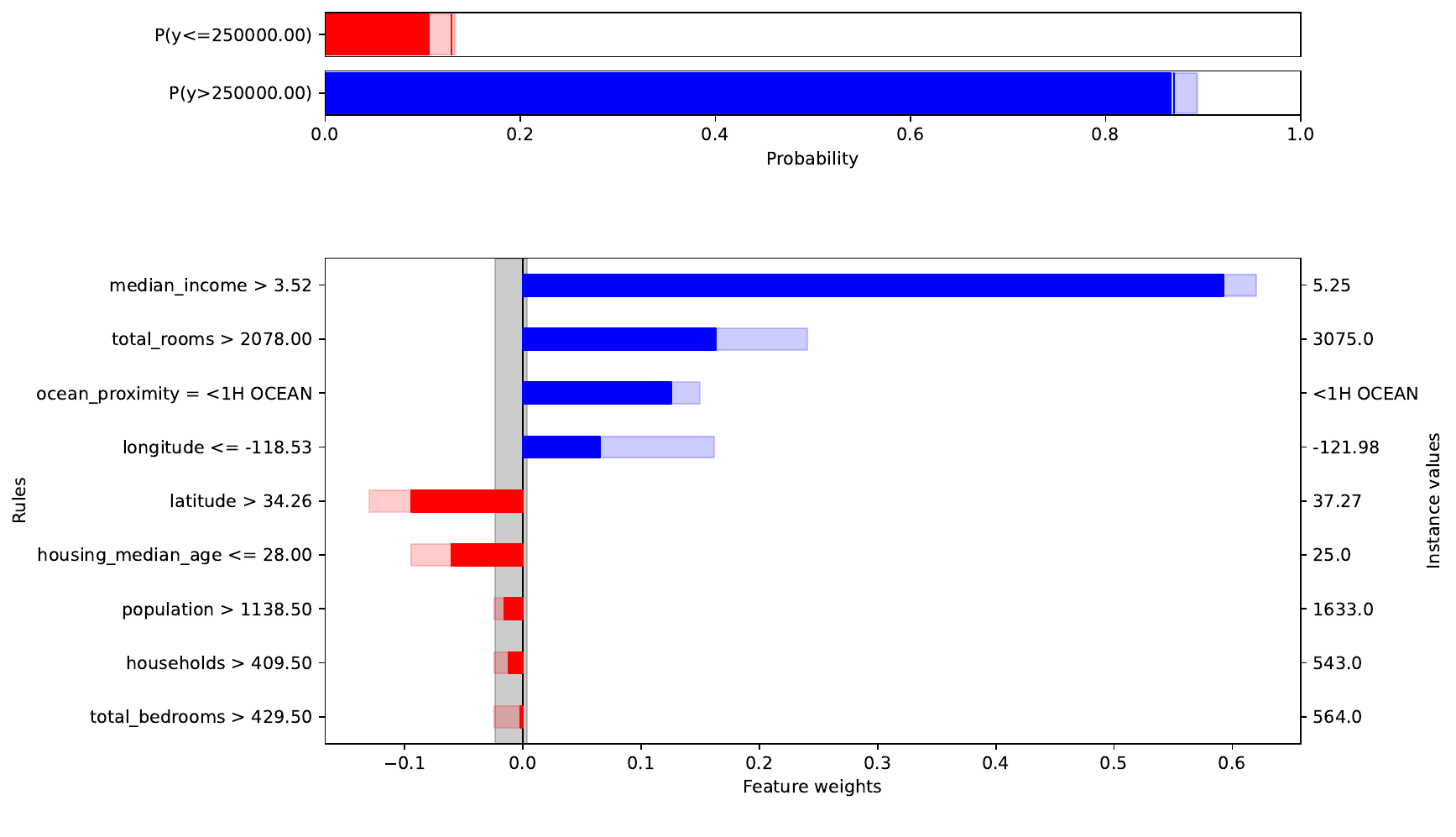}
    \caption{An uncertainty \ri{probabilistic regression plot} for the same explanation as in Fig.~\ref{fig:housing_probability}. The plot includes uncertainties for the feature weights.}
    \label{fig:housing_probability_uncertainty}
\end{figure}

In Fig.~\ref{fig:housing_probability_uncertainty}, the same explanation is shown with uncertainties. As can be seen, the size of the uncertainty varies a lot between features, depending on the calibration of the VA calibrator. 


\subsubsection{Probabilistic Counterfactual Calibrated Explanations for Regression}
\label{sec:CPCER-res}
\begin{figure}[htbp]
    \centering
    \includegraphics[width=0.9\textwidth]{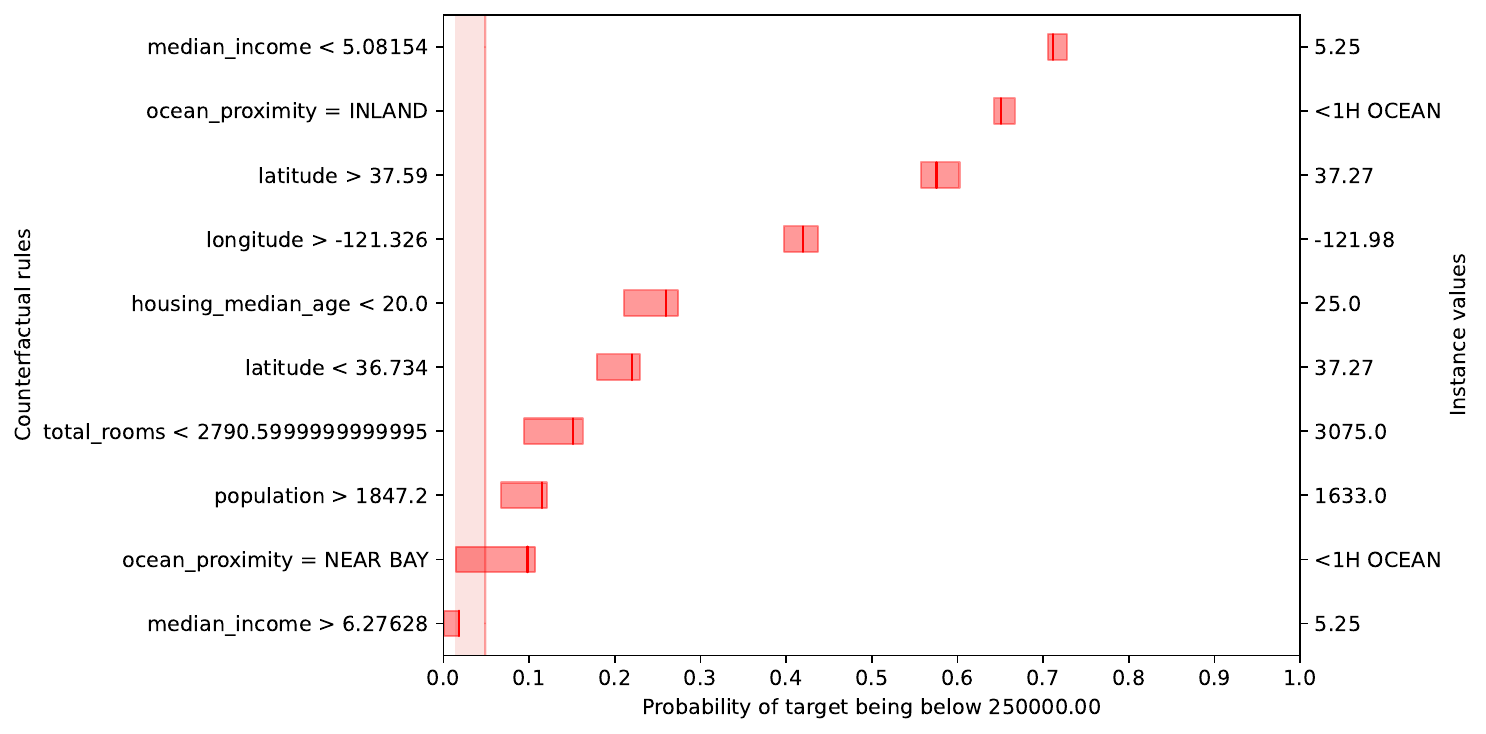}
    \caption{A normalized \ri{probabilistic counterfactual} plot for the same instance as before.}
\label{fig:housing_probability_counter}.
\end{figure}

Fig.~\ref{fig:housing_probability_counter} shows a normalized \ri{probabilistic counterfactual} plot for the same instance. In this case, the normalization used was based on the variance of the predictions of the trees in the random forest. The most influential rule relates to \textit{median income}, with a lower income increasing the probability for a lower price. The normalization will affect the feature probability estimates and confidence intervals and may consequently also result in a different ordering of rules. 

\begin{figure}[H]
    \centering
    \includegraphics[width=0.9\textwidth]{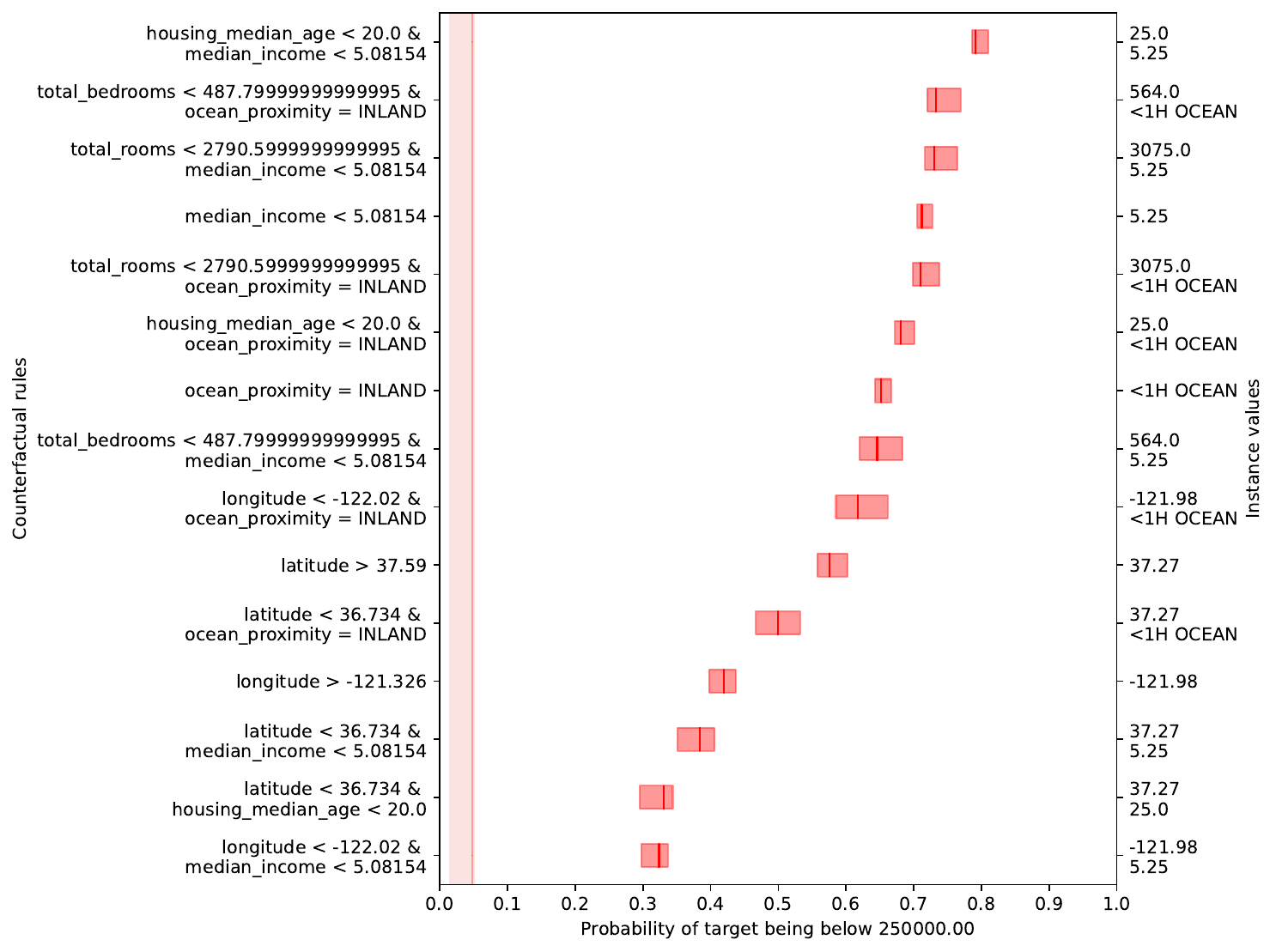}
    \caption{A normalized \ri{probabilistic counterfactual} plot with conjunctive rules for the same instance as before.}
\label{fig:housing_proba_counter_var_conj}.
\end{figure}

The final example, shown in Fig.~\ref{fig:housing_proba_counter_var_conj}, illustrates both conjunctive rules, combining two feature conditions in one rule, and normalization using the variance of the predictions of the trees in the random forest. Here, the number of rules to plot has been increased to 15. Here we see that conjunctive rules often result in more influential rules than single condition rules, illustrated by the majority of rules being conjunctive.

Factual or counterfactual rules can be generated without normalization or with any of the normalization options available in \texttt{DifficultEstimator} in \texttt{crepes.extras}. Conjunctive rules can be added at any time after the explanations are generated. All the examples shown here are from the same instance and the same underlying model, to showcase a subset of available ways the proposed solutions can be used. Further examples can be found in the code repository.

\subsection{Performance Evaluation}
\noindent Table~\ref{tab:performance} shows the results achieved regarding stability, robustness, and \ri{run time}. Stability is measured using the mean variance when constructing explanations on the same instance using different random seeds, with lower values representing more stability. It is evident that both SHAP setups and \ri{both} setups for \ri{standard regression} must be considered stable, since the mean variance is 0 (i.e., less than $1e-31$). LIME and \ri{probabilistic regression}, on the other hand, has a non-negligible mean variance, meaning that they are not, in comparison, as stable. The reason for why \ri{probabilistic regression} is less stable is related to the \ri{sensitivity} of the probabilities derived from the CPD. The reason for the \ri{sensitivity} is that a relatively small change in prediction can easily result in a comparably much larger change in probability for exceeding the threshold, especially if the target is close to the threshold (which is set to $0.5$, i.e., the mid-point in the interval of possible target values). Explanations using the median from a CPD and explanations using the underlying model result in similar stability levels.

Robustness is measured in a similar way as stability, but with a new model trained using different distributions of training and calibration instances between each run. The results achieved on robustness should be seen in relation to the variance in predictions from the underlying model on the same instances. The reason is that if the predictions that the explanations are based on fluctuate, then we can expect a somewhat similar degree of fluctuation in the feature weights as well, since they are defined using the predictions (the mean prediction variance is \ri{$9.1e-5$}). All setups for \ri{Calibrated Explanations} have higher mean variance compared to LIME and SHAP (i.e., are being less robust). However, the explanations produced by the setups for \ri{Calibrated Explanations} do not only rely on the crisp feature weight used to measure the mean variance \ri{(i.e., robustness metric)} but also include the uncertainty interval, highlighting the degree of uncertainty associated with each feature weight. 

\begin{table}[htbp]
    \centering
\begin{tabular}{l|cccccccc}
      & \multicolumn{1}{l}{\textbf{FCER}} & \multicolumn{1}{l}{\textbf{CCER}} & \multicolumn{1}{l}{\textbf{PFCER}} & \multicolumn{1}{l}{\textbf{PCCER}} & \multicolumn{1}{l}{\textbf{LIME}} & \multicolumn{1}{l}{\textbf{LIME}} & \multicolumn{1}{l}{\textbf{SHAP}} & \multicolumn{1}{l}{\textbf{SHAP}} \\
      &       &       &       &       &       & \multicolumn{1}{l}{\textbf{CPS}} &       & \multicolumn{1}{l}{\textbf{CPS}} \\
\hline
Stability  &0 & 0 & 2.2e-3 & 2.7e-3 & 2.7e-5 & 2.7e-5 & 0 & 0 \\
Robustness  & 8.0e-3 & 2.1e-3 & 3.4e-2 & 1.3e-2 & 8.8e-4 & 8.7e-4 & 1.4e-4 & 1.4e-4 \\
\ri{Run} time  & 0.269 & 0.400   & 0.614 & 0.880  & 0.166 & 0.188 & 0.431 & 0.587 \\
\end{tabular}%
    \caption{Evaluation of stability, robustness and \ri{run time}}
    \label{tab:performance}
\end{table}

Regarding \ri{run time}, \ri{all setups have used the same calibration set of 500 instances, including LIME and SHAP. LIME is the fastest and the difference between when explaining the underlying model or when using a CPS is small. Both FCER and CCER are faster than SHAP. The difference between SHAP explaining the underlying model or when using a CPS is fairly large. } 
\ri{PFCER is slightly slower than SHAP CPS}, having to calculate probabilities for \ri{half the} calibration instances as well as training two isotonic calibrators for each test instance. \ri{PCCER is the slowest alternative, having the same overhead as PFCER but also generating a larger number of rules.}

\begin{table}[htbp]
    \centering
    \begin{tabular}{l|cccc|c}
         & \multicolumn{4}{c|}{\textbf{Explanations}} &  \\  
        \textbf{Normalization} & \textbf{FCER} & \textbf{CCER} & \textbf{PFCER} & \textbf{PCCER} & \textbf{Average} \\
        \hline
        \textbf{None} & 0.269 & 0.400 & 0.614 & 0.880 & \textbf{0.541} \\
        \textbf{Distance} & 1.728 & 2.452 & 1.838 & 2.640 & \textbf{2.165} \\
        \textbf{Standard Deviation} & 1.799 & 2.534 & 1.884 & 2.691 & \textbf{2.227} \\
        \textbf{Absolute Error} & 1.849 & 2.613 & 1.954 & 2.757 & \textbf{2.293} \\
        \textbf{Variance} & 1.005 & 1.390 & 1.456 & 1.962 & \textbf{1.453} \\
        \hline
        \textbf{Average} & \textbf{1.330} & \textbf{1.878} & \textbf{1.549} & \textbf{2.186} & \textbf{1.736} \\
    \end{tabular}%
    \caption{\ri{Run time for different kinds of explanations and normalization}}
    \label{tab:comp_time}%
\end{table}%

\ri{Table~\ref{tab:comp_time} show the average time in seconds per instance for creating an explanation with and without normalization for the different kinds of Calibrated Explanations. The most striking result is that using normalization adds a substantial overhead compared to not using normalization: an average of $6\times$ increase in run time. It is also evident that the \textit{k-Nearest Neighbor} based difficulty estimators (using \textit{Distance}, \textit{Standard Deviation} or \textit{Absolute Error} among neighbors) are clearly slower than using the \textit{Variance} among ensemble base regressors. Furthermore, counterfactual explanations are slightly more costly than factual, which is not surprising as they generally generate a larger number of rules. Standard explanations are slightly more than twice as fast as probabilistic explanations without normalization. With normalization, the difference is much smaller, stemming from the fact that only half the calibration set needs normalization, as the other half is used by VA to calibrate the probabilities.}

\ri{Detailed results comparing stability and robustness for different kinds of difficulty estimations is not included, as the differences compared to not using normalization (see Table~\ref{tab:performance}) is small. Detailed results can be found in the \textit{evaluation/regression} folder in the repository.}

\section{Concluding Discussion}
\noindent This paper extends Calibrated Explanations, previously introduced for classification, with support for regression. Two primary use cases are identified: standard regression and probabilistic regression, i.e., measuring the probability of exceeding a threshold. The proposed solution relies on Conformal Predictive Systems (CPS), making it possible to meet the different requirements of the two identified use cases. The proposed solutions provide access to factual and counterfactual explanations with the possibility of conveying uncertainty quantification for the feature rules, just like \ri{Calibrated Explanations for classification}. 

In the paper, the solutions have been demonstrated using several plots, showcasing some of the many ways that the proposed solutions can be used. Furthermore, the paper also includes a comparison with some of the best-known state-of-the-art explanation methods (LIME and SHAP). The results demonstrate that the proposed solution for standard regression is both stable and robust. Furthermore, it is reasonably fast. The suggested solution is considered reliable for two reasons: 1) The calibration of the underlying model and 2) the uncertainty quantification, highlighting the degree of uncertainty of both prediction and feature weights. 

The solution proposed to build probabilistic explanations for regression does not share all the benefits seen for standard regression. The solution has comparable performance as LIME, even if it is clearly slower than LIME. The main strength of this solution is that it provides the possibility of getting probabilistic explanations in relation to an arbitrary threshold from any standard regression model without having to impose any restrictions on the regression model. 

\subsection{Future Work}\label{future-work}
\noindent There are several directions for future work.
An interesting area to look into is how this technique can be adapted to explanations of time-series problems. How to capture and convey the dependency between different time steps pose an interesting challenge.

There are room for improvement regarding \ri{plot layout. Providing} additional ways of visualization is a natural development in the future. This involves implementing support for explanations within image and text prediction, even if these improvements are more closely connected to classification problems.   

\ri{Another direction for future work is to look into probabilistic explanations using the form $\mathcal{P}(t1 < y \leq t2)$. Such predictions would complement the interval predictions provided by CPS by allowing the user to specify the upper and lower bounds of the uncertainty interval and provide the probability of the true target being inside that interval.} 

\ri{Currently, the average calibrated value is used to define the feature weights in equations~(\ref{eq:w_f}), (\ref{eq:w_0^f}), and~(\ref{eq:w_1^f}) (see Section~\ref{CEC}). There are alternatives to taking the average of the perturbed instances for a specific feature and there is room for theoretical analysis on how the feature weights should be calculated to provide the best insights.}

Finally, \ri{run time} can probably be decreased if implementing the core in C++ or by relying on fast languages being able to run Python code more efficiently, e.g., Mojo. 



\backmatter







\section*{Declarations}


\begin{itemize}
\item Funding: The authors acknowledge the Swedish Knowledge Foundation and industrial partners for financially supporting the research and education environment on Knowledge Intensive Product Realization SPARK at Jönköping University, Sweden. Projects: AFAIR grant no. 20200223, ETIAI grant no. 20230040, and PREMACOP grant no. 20220187. Helena Löfström was a PhD student in the Industrial Graduate School in Digital Retailing (INSiDR) at the University of Borås, funded by the Swedish Knowledge Foundation, grant no. 20160035 when this work was drafted. 
\item Competing interests: None
\item Ethics approval: Not applicable.
\item Consent to participate: Not applicable, no respondents involved.
\item Consent for publication: All authors consent to publication. No respondents were involved. 
\item Availability of data and materials: \ri{The \textit{evaluation/regression} folder in the repository contains code for reproducing experiments. Plots corresponding to the once included in the paper can be created using the demos for regression and probabilistic regression in the \textit{notebooks} folder.}
\item Code availability:  
\href{https://github.com/Moffran/calibrated_explanations} {github.com$/$Moffran$/$calibrated$\_$explanations} contains the Python package \texttt{calibrated-explanations}
which is also available for installation through \texttt{pip install calibrated-explanations} or through \texttt{conda install -c conda-forge calibrated-explanations}. 
\item Authors' contributions: Tuwe Löfström has implemented both the extensions for regression and the experiments and has written the major part of the paper. Helena Löfström is the original inventor of the Calibrated Explanations and has taken an active part in both discussions and in writing primarily the introduction and background. Ulf Johansson and Cecilia Sönströd have contributed actively to discussions. Ulf Johansson also contributed with an important improvement making it possible to get rid of a direct reliance on lime. Cecilia Sönströd has also been proofreading. Rudy Matela worked on the workflow for packaging and release of the implementation as well as helping out with proofreading the paper.
\end{itemize}

\bibliography{sample}

\end{document}